\documentclass[11pt]{article}
\usepackage[]{acl}

\usepackage[T1]{fontenc}

\usepackage{times}
\usepackage{latexsym}
\usepackage{microtype}
\usepackage{inconsolata}
\usepackage{amsmath}

\usepackage{booktabs}
\usepackage{colortbl}
\usepackage{multirow}
\usepackage{arydshln}
\usepackage{ragged2e}
\usepackage{xcolor}

\usepackage{arabtex}
\usepackage{utf8}

\usepackage{graphicx}
\usepackage{cuted}
\usepackage{caption}
\usepackage{subcaption}

\setcode{utf8}

\makeatletter
\newcommand{\midsize}{\fontsize{9.5pt}{11.4pt}\selectfont} 

\renewenvironment{abstract}{%
  \begin{center}
    {\bfseries Abstract}
  \end{center}
  \quotation           
  \midsize             
}{%
  \endquotation
  \par\vspace{1em}
}
\makeatother

\title{Zero-Shot Context-Aware ASR for Diverse Arabic Varieties}

\author{
Bashar Talafha \\
The University of British Columbia \\
  {\tt btalafha@mail.ubc.ca}
  \And
Amin Abu Alhassan \\
Imperial College London \\
  {\tt amin.abualhassan25@imperial.ac.uk}
  \AND
Muhammad Abdul-Mageed \\
Canada Research Chair in NLP and ML\\The University of British Columbia\\
  {\tt muhammad.mageed@ubc.ca}
}

\begin{document}
\maketitle
\begin{strip}
\captionsetup{type=figure} 
\centering

\begin{subfigure}[b]{0.31\textwidth}
  \includegraphics[width=\textwidth]{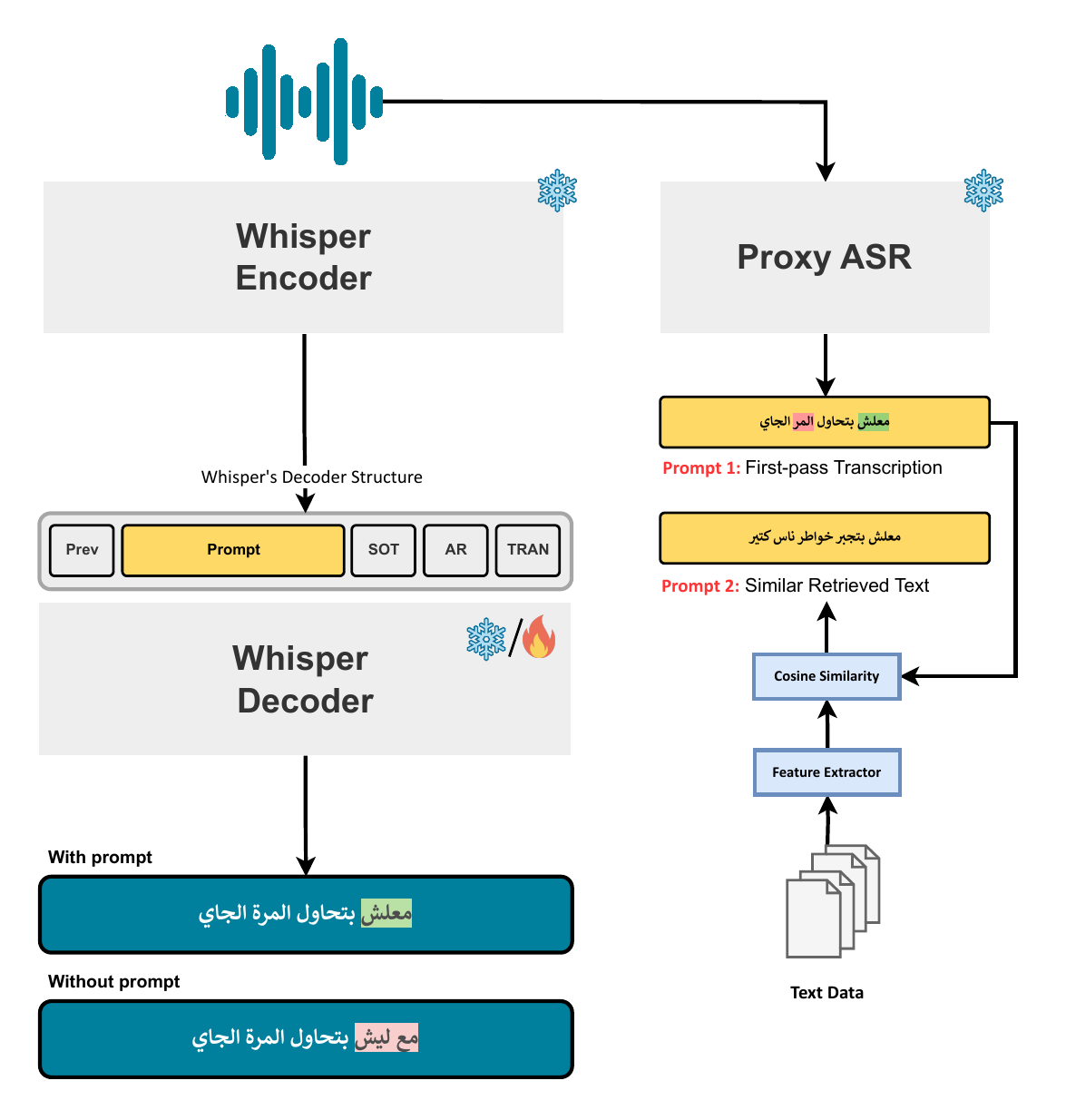}
  \caption{\textbf{Prompt-based context integration}. Whisper’s decoder is conditioned on contextual text via (\textcolor{red}{\texttt{Prompt1}}) a prior transcription or (\textcolor{red}{\texttt{Prompt2}}) a retrieved similar sentence, improving zero-shot dialectal transcription.}
  \label{fig:prompt}
\end{subfigure}
\hfill
\begin{subfigure}[b]{0.32\textwidth}
  \includegraphics[width=\textwidth]{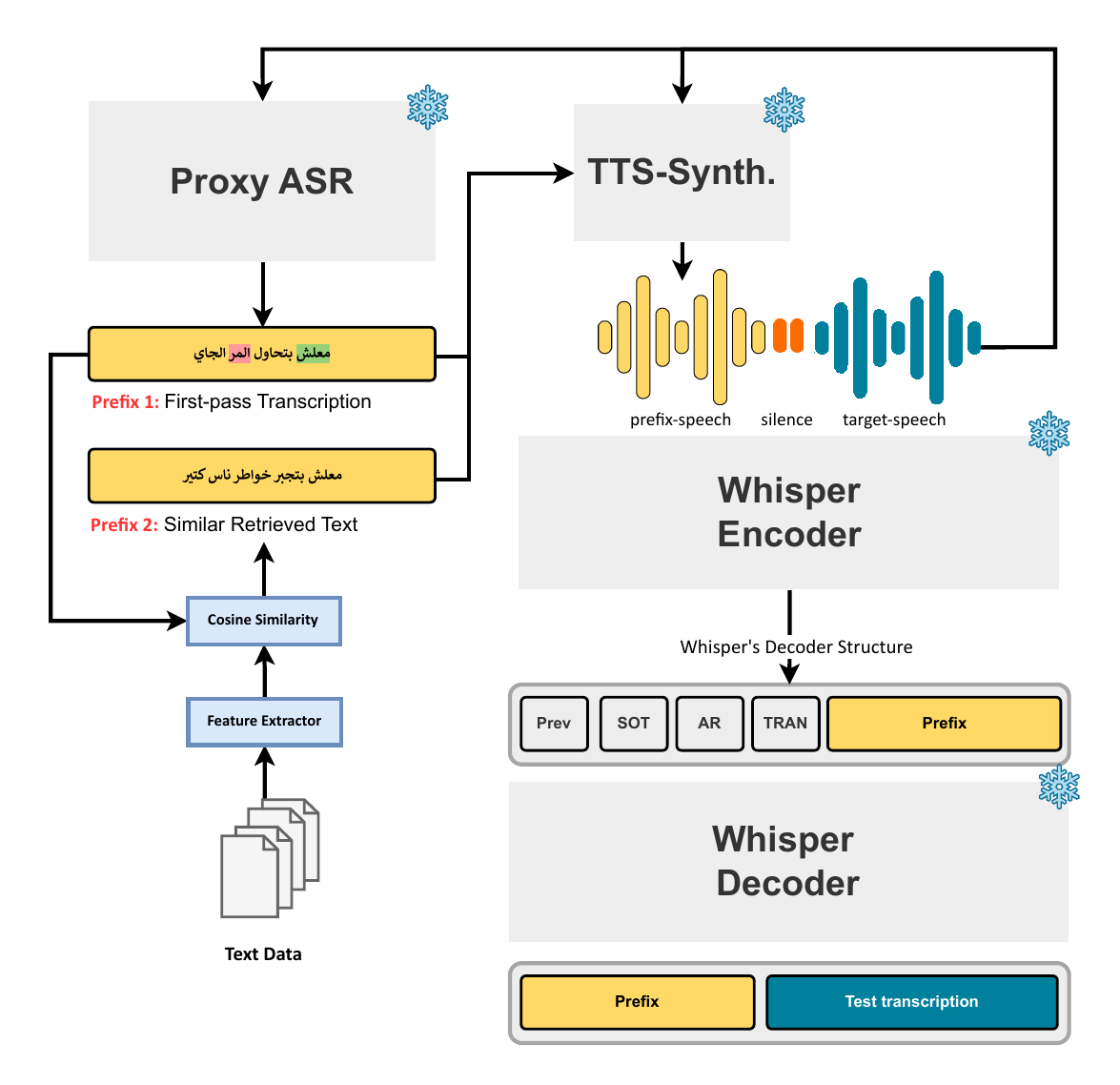}
  \caption{\textbf{Prefix-based context integration}. We retrieve similar (audio, text) pairs and prepend the text (\textcolor{red}{\texttt{Prefix}}) to the decoder and the corresponding retrieved or speaker-conditioned synthesized audio to the encoder.}
  \label{fig:prefix}
\end{subfigure}
\hfill
\begin{subfigure}[b]{0.35\textwidth}
  \includegraphics[width=\textwidth]{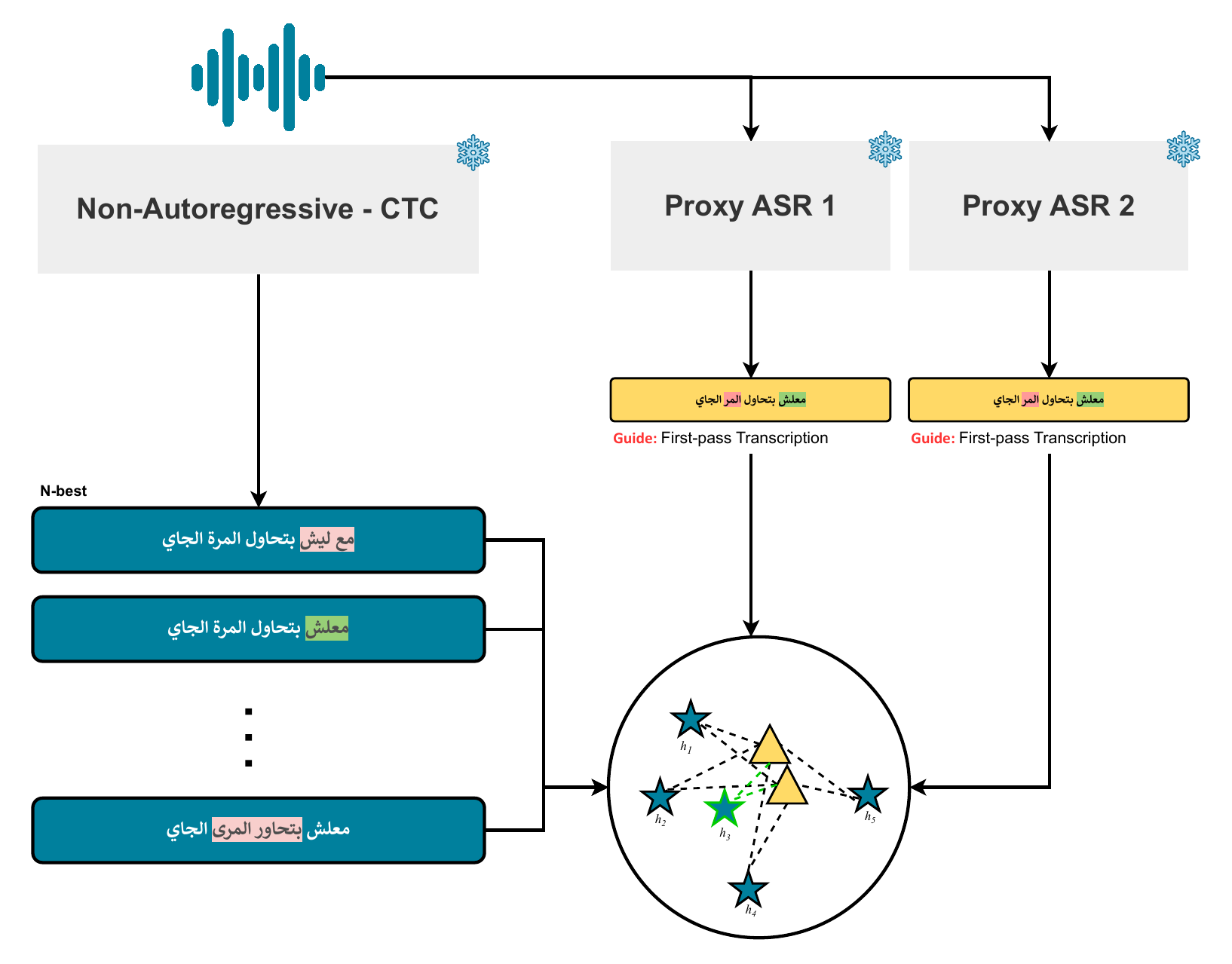}
  \caption{\textbf{Proxy-guided n-best selection}. One or more proxy transcriptions \textcolor{red}{\texttt{guide}} the selection of the closest hypothesis from a non-autoregressive ASR model’s n-best list under WER, consistently outperforming top-1 decoding and approaching the oracle.}
  \label{fig:ctc}
\end{subfigure}

\caption{Zero-shot context-aware adaptation strategies in this work.}

\label{fig:context-aware}
\end{strip}

\begin{abstract}
Zero-shot ASR for Arabic remains challenging: while multilingual models perform well on Modern Standard Arabic (MSA), error rates rise sharply on dialectal and accented speech due to linguistic mismatch and scarce labeled data. We study \textit{context-aware decoding} as a lightweight test-time adaptation paradigm that conditions inference on external side information without parameter updates. For promptable encoder–decoder ASR (e.g., Whisper), we incorporate context through (i) decoder prompting with first-pass hypotheses and (ii) encoder/decoder prefixing with retrieved speech-text exemplars, complemented by simple prompt reordering and optional speaker-matched synthetic exemplars to improve robustness in informal and multi-speaker settings. To extend contextual adaptation beyond promptable architectures, we introduce \textit{proxy-guided $n$-best selection} for CTC ASR: given one or more external proxy hypotheses, we select from a model's $n$-best list by minimizing text-level distance to the proxies, enabling contextual inference without direct prompting. Across ten Arabic conditions spanning MSA, accented MSA, and multiple dialects, context-aware decoding yields average relative WER reductions of 22.29\% on MSA, 20.54 on accented MSA, and 9.15\% on dialectal Arabic. For CTC models, proxy-guided selection reduces WER by 15.6\% relative on MSA and recovers a substantial fraction of oracle $n$-best gains, demonstrating that context-aware inference generalizes beyond encoder-decoder ASR.
\end{abstract}

\section{Introduction}

Large-scale multilingual ASR has substantially improved recognition for high-resource languages~\cite{sekoyan2025canary,pratap2023,babu2021}. Models such as Whisper~\cite{radford2023robust} achieve strong performance on languages including English and Modern Standard Arabic (MSA)~\cite{abdelali2023larabench}. However, performance on dialectal Arabic remains markedly lower~\cite{team2025fanar,talafha2024casablanca,talafha2023n}, reflecting phonological, lexical, and syntactic differences from MSA~\cite{ali2016msa_dialect}, as well as orthographic variability, code-switching, and persistent data mismatch. Since collecting sufficient labeled speech for each variety is costly and often infeasible~\cite{jm3}, \textit{zero-shot} ASR (i.e., inference without dialect-specific supervised adaptation or parameter updates) is a practical necessity. Consequently, dialectal Arabic continues to exhibit high error rates on benchmarks such as Casablanca~\cite{talafha2024casablanca}.

In this work, we explore \textit{context-aware decoding} as a lightweight test-time adaptation strategy for zero-shot Arabic ASR. The core idea is to condition decoding on external side information available at inference time, such as first-pass hypothesis or retrieved in-domain exemplars, to steer recognition without retraining. We instantiate this in Whisper, a promptable encoder–decoder model, by injecting contextual text as decoder prompts and by prefixing speech–text exemplars at the encoder and decoder inputs. We further study simple prompt reordering and speaker-matched synthetic exemplars to improve robustness in noisy, multi-speaker, or conversational settings.

Importantly, context-aware inference need not rely on promptable architectures. To extend the same principle to non-autoregressive ASR,  we propose \textit{proxy-guided $n$-best selection} for encoder-only CTC models, such as Omnilingual ASR~\cite{omnilingual2025omnilingual}. In this setting, one or more external ASR hypotheses act as \textit{proxies} that guide selection from the model's $n$-best list using text-level distance metrics. This yields a simple, training-free mechanism for contextual adaptation even when direct prompting is unavailable.

We evaluate our methods on ten Arabic conditions spanning MSA, accented MSA, and multiple regional dialects. Across conditions, context-aware decoding improves over Whisper and SeamlessM4T~\cite{barrault2023seamlessm4t} baselines, yielding average relative WER reductions of 22.29\% on MSA, 20.54\% on accented MSA, and 9.15\% on dialectal Arabic. Proxy-guided $n$-best selection provides additional gains for CTC ASR, reducing WER by 15.6\% relative on MSA and recovering a substantial fraction of oracle $n$-best improvements. Our results suggest that context-aware inference is a general and practical approach to improving zero-shot ASR in dialect-rich, low-resource settings.


\section{Related Work}

The persistent performance gap between ASR systems on high-resource languages and low-resource dialects has motivated a wide range of adaptation strategies. Prior work has shown that even state-of-the-art multilingual models such as Whisper and SeamlessM4T perform poorly in zero-shot dialectal settings~\cite{talafha2024casablanca,abdelali2023larabench}. In particular, Whisper is prone to hallucinated, repetitive, or boilerplate outputs when decoding speech from unseen dialects or informal domains~\cite{talafha2023n}. Approaches based on model distillation, such as uDistil-Whisper~\cite{waheed2024udistil}, attempt to address these issues but remain limited by their reliance on pseudo-labels generated by teacher models that themselves underperform on dialectal speech.

More recently, test-time adaptation through contextual prompting has emerged as a promising alternative to fine-tuning. \newcite{suh2024} demonstrate that injecting manually written or automatically generated textual prompts into Whisper’s decoder can significantly improve domain-specific transcription. Complementarily, \newcite{wang2024icl} propose Speech-based In-Context Learning (SICL), which adapts Whisper at inference time by concatenating retrieved speech examples to the encoder input and prepending their transcripts as decoder prefixes, achieving large relative WER reductions on unseen Chinese dialects. These methods enable adaptation without gradient updates and are particularly attractive for low-resource and multilingual settings. Our work builds on this line of research by extending prompt- and prefix-based contextualization to Arabic dialectal ASR and by introducing additional mechanisms—such as prompt reordering, modality-specific retrieval, and speaker-conditioned synthesis—to improve robustness in multi-speaker and informal scenarios.

Beyond prompting, decoding-time alternatives to standard beam search have also been explored. Sample-based Minimum Bayes Risk (MBR) decoding selects hypotheses that maximize agreement within an $n$-best list and has been shown to outperform beam search for autoregressive ASR models such as Whisper~\cite{jinnai2025re}. However, MBR relies exclusively on internal hypothesis structure and does not incorporate external contextual signals. Moreover, its applicability to non-autoregressive encoder-only architectures remains limited. Relatedly, \newcite{cheng2024context} investigate confidence- and similarity-based hypothesis selection strategies at decoding time, highlighting the potential of decision-level interventions without retraining. These approaches, however, do not exploit external proxy transcriptions as a source of contextual guidance.

Our work builds on the general insight that test-time contextual information can meaningfully improve ASR, while addressing limitations of prior methods. In addition to prompt- and prefix-based adaptation for encoder–decoder models, we introduce proxy-guided $n$-best selection for non-autoregressive CTC ASR systems, such as Omnilingual ASR. By leveraging external ASR hypotheses as proxies and selecting the closest candidate under text-level distance metrics, our approach provides a simple yet effective decision-level adaptation mechanism that complements existing prompting and reranking techniques, and extends \textit{context-aware} decoding to architectures that do not support textual prompts.
\section{Methodology}
\label{sec:methodology}

We study \textit{context-aware decoding} as a training-free  test-time adaptation paradigm for ASR. Given an input utterance, we obtain auxiliary \textit{context} from sources available at inference time (e.g., first-pass hypotheses or retrieved exemplars) and incorporate it into decoding \textit{without} updating model parameters. As summarized in Figure~\ref{fig:context-aware}, we instantiate context-aware decoding through three complementary mechanisms, depending on where context enters the inference pipeline: (i) \textbf{decoder prompting} (\S~\ref{sec:prompt}) for promptable encoder–decoder ASR, (ii) \textbf{encoder-decoder prefixing} (\S~\ref{sec:prefix-methods}) with aligned speech–text exemplars, and (ii) \textbf{decision-level reranking} (\S~\ref{sec:ctc}) for non-autoregressive CTC ASR via proxy-guided $n$-best selection.

Throughout, we use Whisper to instantiate prompt- and prefix-based integration, and we extend the same principle to non-promptable architectures by applying proxy-guided selection to Omnilingual. Unless otherwise stated, all retrieval indices are fixed before evaluation and do not include ground-truth transcripts from evaluation sets.

\subsection{Prompt-Based Context Integration}\label{sec:prompt}

Prompt-based context integration leverages the ability of promptable encoder–decoder ASR models to condition decoding on external \textit{textual} side information. We instantiate this strategy using Whisper~\cite{radford2023robust}, a multilingual encoder–decoder Transformer whose autoregressive decoder accepts optional prompt region (i.e., a prefix of text tokens) that biases generation without modifying model parameters.

As illustrated in Figure~\ref{fig:prompt}, Whisper decodes audio conditioned on a structured token sequence consisting of (i) a prompt region (if provided), (ii) by task-specific tokens (language and transcription mode), and (iii) and the output sequence. We inject contextual text immediately after the designated prompt region (i.e., \texttt{|PREV|} token), so that lexical and semantic cues can influence decoding while preserving the standard Whisper inference pipeline. We consider two sources of textual context: (i) a \textit{first-pass} transcription of the target utterance produced by an auxiliary ASR system, and (ii) a \textit{retrieved} sentence selected from a fixed reference text index using the first-pass hypothesis as a query (Figure~\ref{fig:prompt}, Prompt1 and Prompt2). Both variants aim to bias decoding toward lexical choices that better match the target variety without supervised adaptation.

\paragraph{First-pass transcription prompting.}
We use first-pass hypotheses generated by SeamlessM4T (SM4T)~\cite{barrault2023seamlessm4t} as prompts, motivated by its strong zero-shot performance on some Arabic dialects~\cite{waheed2024udistil}. The intuition is that exposing Whisper to dialectal lexical forms at inference time  can bias the decoder toward outputs that better reflect the target variety, even when the first-pass output is imperfect. Because Whisper’s autoregressive decoder may treat prompts as coherent text to be continued, we also explore simple \textbf{prompt reordering} heuristics intended to reduce prompt continuation behavior while preserving lexical content. Concretely, given a prompt token sequence, we apply either \textbf{(a)} random permutation of \textit{word tokens} or \textbf{(b)} reversal at the word level. These transformations preserve the multiset of prompt words but disrupt sequential coherence. Their empirical impact, including effects on hallucination, is analyzed in Section~\ref{sec:context-aware-decoding}.

\paragraph{Retrieval-based prompting.}
To reduce sensitivity to noisy first-pass transcriptions, we alternatively retrieve a similar sentence from a fixed human-written text index and use it as the prompt. Retrieval operates purely in the text domain: given a first-pass hypothesis, we compute its embedding using a shared feature extractor and retrieve the nearest neighbor sentence under cosine similarity. In our experiments, the reference index is constructed from the \textit{text side} of a corpus containing approximately 500K speech–text pairs~\cite{waheed2024udistil}. This setup is attractive in low-resource settings because it relies only on text at inference time and does not require paired contextual audio. 

\subsection{Prefix-Based Context Integration}
\label{sec:prefix-methods}

Prefix-based context integration extends prompt-based conditioning by injecting context into \textit{both} the encoder and decoder of Whisper using aligned speech-text exemplars. Rather than providing text alone, this method prepends a contextual audio segment and its transcript to the test utterance, enabling the model to exploit complementary acoustic and linguistic cues throughout the full encoder–decoder pipeline. Our approach is inspired by SICL~\cite{wang2024icl}, which demonstrates effective test-time conditioning without parameter updates.

As illustrated in Figure~\ref{fig:prefix}, given a test utterance with audio input $\mathbf{x}$, we construct a contextual prefix consisting of a contextual audio $\mathbf{x}_{\mathrm{ctx}}$ and its transcript $\mathbf{p}_{\mathrm{ctx}}$. The concatenated waveform $\mathbf{x}_{\mathrm{ctx}} \oplus \mathbf{x}$, separated by a fixed 1-second silence, is fed to Whisper's encoder. On the decoder side, we prepended $\mathbf{p}_{\mathrm{ctx}}$ to the decoder token history, and then generate the target transcript autoregressively:

\begin{equation*}
\hat{\mathbf{y}}
= \arg\max_{\mathbf{y}}
\prod_{t=1}^{T}
P\!\left(y_t \mid [\mathbf{p}_{\mathrm{ctx}};\mathbf{y}_{<t}], \, \mathbf{x}_{\mathrm{ctx}} \oplus \mathbf{x};\, \theta \right).
\label{eq:prefix}
\end{equation*}

Here $[\mathbf{p}_{\mathrm{ctx}};\mathbf{y}_{<t}]$ denotes concatenation of the fixed contextual transcript prefix and the previously generated tokens. We consider three sources of contextual prefixes. 

\paragraph{Retrieved exemplar prefixing.}
We retrieve a semantically similar utterance from a reference speech-text collection and use its aligned pair $(\mathbf{x}_{\mathrm{ctx}}, \mathbf{p}_{\mathrm{ctx}})$ as context. We retrieve candidates using character-level TF–IDF similarity (Section~\ref{exp-similarity}), which is robust to orthographic variability in dialectal Arabic. Unlike typical SICL setups, retrieved and test utterances in our data may come from different speakers, and we observe that speaker mismatch can destabilize conditioning and degrade decoding quality.\footnote{\url{https://github.com/openai/whisper/discussions/434}}

\paragraph{Speaker-conditioned synthesis for prefix audio.}
To mitigate speaker mismatch, we synthesize the contextual audio to match the target speaker while keeping the contextual text fixed. Specifically, given a contextual transcript $\mathbf{p}_{\mathrm{ctx}}$, we generate speaker-conditioned speech $\tilde{\mathbf{x}}_{\mathrm{ctx}}$ using a pretrained TTS model (XTTS~\cite{casanova2024xtts}) conditioned on a speaker embedding extracted from the test utterance. Because the contextual waveform is generated from text, this variant avoids requiring paired contextual speech at inference time (beyond the test utterance itself) while providing a single-speaker prefix that reduces acoustic discontinuities. 

\paragraph{Self-prefixing.} 
Finally, instead of retrieving external exemplars, we optionally construct a contextual prefix directly from the test utterance: we obtain a first-pass hypothesis for the utterance, treat it as $\mathbf{p}_{\mathrm{ctx}}$, and synthesize $\tilde{\mathbf{x}}_{\mathrm{ctx}}$ via the same speaker-synthesis procedure. This self-prefixing variant provides speaker-conditioned TTS procedure. This yields a speaker-consistent contextual prefix without relying on external audio data.

\subsection{Proxy-Guided $n$-Best Selection for Non-Autoregressive CTC ASR}\label{sec:ctc}

The preceding methods exploit the promptable encoder–decoder structure of Whisper. In contrast, many ASR systems are non-autoregressive encoder-only CTC models that do not support textual prompting or encoder-decoder prefixing. To extend context-aware decoding to this setting, we introduce \textit{proxy-guided $n$-best selection}, a decision-level, training-free adaptation strategy that leverages external ASR hypotheses to rerank the model's candidates (Figure~\ref{fig:ctc}).

We instantiate this approach using Omnilingual ASR~\cite{omnilingual2025omnilingual}, a multilingual CTC-based system. Given an audio input $x$, the model produces an $n$-best list of candidate transcriptions $\{h_1,\ldots,h_n\}$ via its standard CTC beam search. In parallel, an auxiliary ASR system (e.g., SM4T) generates a proxy transcription $g$ for the same utterance. We then select the final hypothesis by minimizing a text-level distance between the candidate and the  proxy:
$
\hat{h} = \arg\min_{h_i \in \mathcal{H}} d(h_i, g),
\label{eq:proxy}
$
 where $d(\cdot,\cdot)$ denotes a text-level distance such as WER, CER, or $1{-}$BLEU. This procedure leaves the CTC model,  objective, and beam-search decoding unchanged and exploits proxy information purely as a post-decoding reranker. While we present the single-proxy case for clarity, the method naturally extends to multiple proxies by aggregating distances across proxies (Section~\ref{sec:multi-proxy}).

\section{Experiments}
\label{sec:experiments}

\definecolor{highlight}{HTML}{d6dbf4} 
\definecolor{base}{HTML}{f0f0f0} 
\definecolor{avg}{HTML}{f6f9ef} 

\begin{table*}[h!]
\small
\centering
\resizebox{\textwidth}{!}{%
\begin{tabular}{llllrccclrcl}
\hline
\noalign{\vskip 0.5ex}
\multicolumn{2}{l}{\multirow{2}{*}{\shortstack[l]{Language\\Condition}}}
& \multicolumn{2}{c}{\cellcolor{base}\textbf{Baselines}}
& \multicolumn{5}{c}{\textbf{Ours:} Prompt-based ($\leftarrow$)}
& \multicolumn{3}{c}{\textbf{Ours:} Prefix-based ($\rightarrow$)} \\
\multicolumn{2}{l}{}   
& \cellcolor{base}SM4T
& \cellcolor{base}W-v3
& W$\leftarrow$FPT
& W$\leftarrow$Rand
& W$\leftarrow$Rev
& W$\leftarrow$SMms
& W$\leftarrow$SSea
& \shortstack[r]{W$\rightarrow$SSea\\No-TTS}
& \shortstack[c]{W$\rightarrow$SSea\\TTS}
& \shortstack[l]{W$\rightarrow$Pt\\TTS} \\
\hline
\multicolumn{12}{l}{\cellcolor{highlight}\textbf{MSA}}                                                                                                                                                                                                                                                                                                                                                                                                                                                                                                                                                                                                                                                                                                                               \\ \hline
\noalign{\vskip 0.5ex}
\multicolumn{2}{l}{CV15.0}                                                                       \begin{tabular}[c]{@{}l@{}}\textit{WER/}\\ \textit{CER}\end{tabular} & \cellcolor{base}\begin{tabular}[c]{@{}l@{}}11.12/\\ 3.55\end{tabular}  & \cellcolor{base}\begin{tabular}[c]{@{}l@{}}15.55/\\ 5.06\end{tabular}  & \textbf{\begin{tabular}[c]{@{}r@{}}10.40 /\\ \underline{3.18}\end{tabular}} & \begin{tabular}[c]{@{}c@{}}12.01/\\ 3.93\end{tabular}           & \begin{tabular}[c]{@{}c@{}}12.12 /\\ 3.88\end{tabular}          & \begin{tabular}[c]{@{}c@{}}13.69/\\ 4.59\end{tabular}           & \begin{tabular}[c]{@{}l@{}}12.69/\\ 4.29\end{tabular}           & \begin{tabular}[c]{@{}r@{}}15.67/\\ 7.45\end{tabular}           & \textbf{\begin{tabular}[c]{@{}c@{}}11.26/\\ 3.46\end{tabular}}  & \textbf{\begin{tabular}[c]{@{}l@{}}\underline{10.28}/\\ 3.29\end{tabular}}  \\ 
\noalign{\vskip 0.9ex}

\multicolumn{2}{l}{MGB2}                             \begin{tabular}[c]{@{}l@{}}\textit{WER/}\\ \textit{CER}\end{tabular}                                             & \cellcolor{base}\begin{tabular}[c]{@{}l@{}}17.35/\\ 8.73\end{tabular}  & \cellcolor{base}\begin{tabular}[c]{@{}l@{}}16.02/\\ 7.64\end{tabular}  & \begin{tabular}[c]{@{}r@{}}47.61/\\ 36.61\end{tabular}          & \begin{tabular}[c]{@{}c@{}}16.7/\\ 9.09\end{tabular}            & \textbf{\begin{tabular}[c]{@{}c@{}}15.01/\\ 7.66\end{tabular}}  & \begin{tabular}[c]{@{}c@{}}17.28/\\ 7.48\end{tabular}           & \begin{tabular}[c]{@{}l@{}}16.51/\\ 7.24\end{tabular}           & \begin{tabular}[c]{@{}r@{}}17.15/\\ 7.94\end{tabular}           & \textbf{\begin{tabular}[c]{@{}c@{}}14.66/\\ \underline{5.97}\end{tabular}}  & \textbf{\begin{tabular}[c]{@{}l@{}}\underline{14.26}/\\ 6.36\end{tabular}}  \\
\noalign{\vskip 0.9ex}
\multicolumn{2}{l}{\textit{Avg MSA}}  \begin{tabular}[c]{@{}l@{}}\textit{WER/}\\ \textit{CER}\end{tabular}                                                           & \cellcolor{base}\begin{tabular}[c]{@{}l@{}}14.24/\\ 06.14\end{tabular} & \cellcolor{base}\begin{tabular}[c]{@{}l@{}}15.79/\\ 06.35\end{tabular} & \begin{tabular}[c]{@{}r@{}}29.01/\\ 19.90\end{tabular}          & \begin{tabular}[c]{@{}c@{}}14.36/\\ 06.51\end{tabular}          & \textbf{\begin{tabular}[c]{@{}c@{}}13.57/\\ 05.77\end{tabular}} & \textbf{\begin{tabular}[c]{@{}c@{}}15.49/\\ 06.04\end{tabular}} & \textbf{\begin{tabular}[c]{@{}l@{}}14.60/\\ 05.77\end{tabular}} & \begin{tabular}[c]{@{}r@{}}16.41/\\ 07.70\end{tabular}          & \textbf{\begin{tabular}[c]{@{}c@{}}12.96/\\ \underline{04.72}\end{tabular}} & \textbf{\begin{tabular}[c]{@{}l@{}}\underline{12.27}/\\ 04.83\end{tabular}} \\ \hline
\multicolumn{12}{l}{\cellcolor{highlight}\textbf{Accented MSA}}                                                                                                                                                                                                                                                                                                                                                                                                                                                                                                                                                                                                                                                                                                                          \\ \hline
\noalign{\vskip 0.5ex}
\multicolumn{2}{l}{Fleurs}  \begin{tabular}[c]{@{}l@{}}\textit{WER/}\\ \textit{CER}\end{tabular}                                                                      &\cellcolor{base} \begin{tabular}[c]{@{}l@{}}7.66/\\ 4.0\end{tabular}    &\cellcolor{base} \begin{tabular}[c]{@{}l@{}}9.2/\\ 2.77\end{tabular}    & \begin{tabular}[c]{@{}r@{}}17.34/\\ 12.56\end{tabular}          & \textbf{\begin{tabular}[c]{@{}c@{}}7.36/\\ 3.73\end{tabular}}   & \textbf{\begin{tabular}[c]{@{}c@{}}\underline{7.31}/\\ 3.76\end{tabular}}   & \begin{tabular}[c]{@{}c@{}}12.18/\\ 3.93\end{tabular}           & \begin{tabular}[c]{@{}l@{}}12.21/\\ 4.34\end{tabular}           & \begin{tabular}[c]{@{}r@{}}11.56/\\ 4.69\end{tabular}           & \begin{tabular}[c]{@{}c@{}}10.22/\\ 3.60\end{tabular}           & \begin{tabular}[c]{@{}l@{}}9.31/\\ \textbf{\underline{2.72}}\end{tabular}            \\ \hline
\multicolumn{12}{l}{\cellcolor{highlight}\textbf{External Dialects}}                                                                                                                                                                                                                                                                                                                                                                                                                                                                                                                                                                                                                                                                                                                     \\ \hline
\noalign{\vskip 0.5ex}
\multicolumn{2}{l}{MGB3}    \begin{tabular}[c]{@{}l@{}}\textit{WER/}\\ \textit{CER}\end{tabular}                                                                & \cellcolor{base}\begin{tabular}[c]{@{}l@{}}31.48/\\ 15.75\end{tabular} &\cellcolor{base} \begin{tabular}[c]{@{}l@{}}35.9/\\ 17.67\end{tabular}  & \begin{tabular}[c]{@{}r@{}}63.47/\\ 50.14\end{tabular}          & \begin{tabular}[c]{@{}c@{}}31.62/\\ 15.98\end{tabular}          & \textbf{\begin{tabular}[c]{@{}c@{}}\underline{31.14}/\\ \underline{15.26}\end{tabular}} & \begin{tabular}[c]{@{}c@{}}37.15/\\ 18.43\end{tabular}          & \begin{tabular}[c]{@{}l@{}}35.45/\\ 17.47\end{tabular}          & \begin{tabular}[c]{@{}r@{}}35.45/\\ 17.22\end{tabular}          & \begin{tabular}[c]{@{}c@{}}34.22/\\ 15.90\end{tabular}          & \begin{tabular}[c]{@{}l@{}}33.51/\\ 18.64\end{tabular}          \\
\noalign{\vskip 0.9ex}
\multicolumn{2}{l}{MGB5}    \begin{tabular}[c]{@{}l@{}}\textit{WER/}\\ \textit{CER}\end{tabular}                                                                & \cellcolor{base}\begin{tabular}[c]{@{}l@{}}77.43/\\ 43.62\end{tabular} & \cellcolor{base}\begin{tabular}[c]{@{}l@{}}79.16/\\ 45.1\end{tabular}  & \textbf{\begin{tabular}[c]{@{}r@{}}76.04/\\ 45.66\end{tabular}} & \textbf{\begin{tabular}[c]{@{}c@{}}69.37/\\ 35.55\end{tabular}} & \textbf{\begin{tabular}[c]{@{}c@{}}69.89/\\ 35.49\end{tabular}} & \textbf{\begin{tabular}[c]{@{}c@{}}76.90/\\ 42.02\end{tabular}} & \textbf{\begin{tabular}[c]{@{}l@{}}75.23/\\ 40.37\end{tabular}} & \textbf{\begin{tabular}[c]{@{}r@{}}74.35/\\ 38.90\end{tabular}} & \textbf{\begin{tabular}[c]{@{}c@{}}73.70/\\ 37.13\end{tabular}} & \textbf{\begin{tabular}[c]{@{}l@{}}\underline{68.21}/\\ \underline{33.96}\end{tabular}} \\
\noalign{\vskip 0.9ex}
\multicolumn{2}{l}{\textit{Avg Ext}}   \begin{tabular}[c]{@{}l@{}}\textit{WER/}\\ \textit{CER}\end{tabular}                                                           & \cellcolor{base}\begin{tabular}[c]{@{}l@{}}54.46/\\ 29.69\end{tabular} & \cellcolor{base}\begin{tabular}[c]{@{}l@{}}57.53/\\ 31.39\end{tabular} & \begin{tabular}[c]{@{}r@{}}69.76/\\ 47.90\end{tabular}          & \textbf{\begin{tabular}[c]{@{}c@{}}50.50/\\ 25.77\end{tabular}} & \textbf{\begin{tabular}[c]{@{}c@{}}\underline{50.52}/\\ \underline{25.38}\end{tabular}} & \begin{tabular}[c]{@{}c@{}}57.03/\\ 30.23\end{tabular}          & \begin{tabular}[c]{@{}l@{}}55.34/\\ 28.92\end{tabular}          & \textbf{\begin{tabular}[c]{@{}r@{}}54.90/\\ 28.06\end{tabular}} & \textbf{\begin{tabular}[c]{@{}c@{}}53.96/\\ 26.52\end{tabular}} & \textbf{\begin{tabular}[c]{@{}l@{}}50.86/\\ 26.30\end{tabular}} \\ \hline
\multicolumn{12}{l}{\cellcolor{highlight}\textbf{Casa-\textit{Dialects}}}                                                                                                                                                                                                                                                                                                                                                                                                                                                                                                                                                                                                                                                                                                                     \\ \hline
\noalign{\vskip 0.5ex}
\multicolumn{2}{l}{ALG}   \begin{tabular}[c]{@{}l@{}}\textit{WER/}\\ \textit{CER}\end{tabular}                                                                        & \cellcolor{base}\begin{tabular}[c]{@{}l@{}}86.89/\\ 45.5\end{tabular}  & \cellcolor{base}\begin{tabular}[c]{@{}l@{}}78.6/\\ 37.81\end{tabular}  & \textbf{\begin{tabular}[c]{@{}r@{}}77.83/\\ 39.19\end{tabular}} & \textbf{\begin{tabular}[c]{@{}c@{}}73.07/\\ 31.26\end{tabular}} & \textbf{\begin{tabular}[c]{@{}c@{}}73.13/\\ \underline{30.38}\end{tabular}} & \textbf{\begin{tabular}[c]{@{}c@{}}76.59/\\ 34.70\end{tabular}} & \textbf{\begin{tabular}[c]{@{}l@{}}74.68/\\ 34.17\end{tabular}} & \textbf{\begin{tabular}[c]{@{}r@{}}76.87/\\ 36.57\end{tabular}} & \textbf{\begin{tabular}[c]{@{}c@{}}74.53/\\ 31.28\end{tabular}} & \textbf{\begin{tabular}[c]{@{}l@{}}\underline{70.08}/\\ 31.85\end{tabular}} \\
\noalign{\vskip 0.9ex}
\multicolumn{2}{l}{JOR}    \begin{tabular}[c]{@{}l@{}}\textit{WER/}\\ \textit{CER}\end{tabular}                                                                       & \cellcolor{base}\begin{tabular}[c]{@{}l@{}}38.29/\\ 12.01\end{tabular} &\cellcolor{base} \begin{tabular}[c]{@{}l@{}}40.79/\\ 13.55\end{tabular} & \textbf{\begin{tabular}[c]{@{}r@{}}37.34/\\ 12.12\end{tabular}} & \textbf{\begin{tabular}[c]{@{}c@{}}37.52/\\ 12.25\end{tabular}} & \textbf{\begin{tabular}[c]{@{}c@{}}37.34/\\ 12.12\end{tabular}} & \textbf{\begin{tabular}[c]{@{}c@{}}38.01/\\ 13.39\end{tabular}} & \textbf{\begin{tabular}[c]{@{}l@{}}35.90/\\ 12.59\end{tabular}} & \textbf{\begin{tabular}[c]{@{}r@{}}36.96/\\ 13.61\end{tabular}} & \textbf{\begin{tabular}[c]{@{}c@{}}\underline{35.00}/\\ \underline{11.79}\end{tabular}} & \textbf{\begin{tabular}[c]{@{}l@{}}36.04/\\ 14.10\end{tabular}} \\
\noalign{\vskip 0.9ex}
\multicolumn{2}{l}{PAL}     \begin{tabular}[c]{@{}l@{}}\textit{WER/}\\ \textit{CER}\end{tabular}                                                                      & \cellcolor{base}\begin{tabular}[c]{@{}l@{}}48.82/\\ 16.49\end{tabular} &\cellcolor{base} \begin{tabular}[c]{@{}l@{}}50.38/\\ 17.52\end{tabular} & \textbf{\begin{tabular}[c]{@{}r@{}}46.12/\\ 14.98\end{tabular}} & \textbf{\begin{tabular}[c]{@{}c@{}}46.55/\\ \underline{14.9}\end{tabular}}  & \textbf{\begin{tabular}[c]{@{}c@{}}46.12/\\ 14.96\end{tabular}} & \textbf{\begin{tabular}[c]{@{}c@{}}45.28/\\ 16.48\end{tabular}} & \textbf{\begin{tabular}[c]{@{}l@{}}45.24/\\ 16.69\end{tabular}} & \textbf{\begin{tabular}[c]{@{}r@{}}44.38/\\ 16.65\end{tabular}} & \textbf{\begin{tabular}[c]{@{}c@{}}\underline{42.94}/\\ 14.92\end{tabular}} & \textbf{\begin{tabular}[c]{@{}l@{}}52.78/\\ 27.16\end{tabular}} \\
\noalign{\vskip 0.9ex}
\multicolumn{2}{l}{UAE}     \begin{tabular}[c]{@{}l@{}}\textit{WER/}\\ \textit{CER}\end{tabular}                                                                      &\cellcolor{base} \begin{tabular}[c]{@{}l@{}}51.79/\\ 19.75\end{tabular} &\cellcolor{base} \begin{tabular}[c]{@{}l@{}}55.03/\\ 22.98\end{tabular} & \textbf{\begin{tabular}[c]{@{}r@{}}49.1/\\ 18.13\end{tabular}}  & \textbf{\begin{tabular}[c]{@{}c@{}}49.45/\\ 18.48\end{tabular}} & \textbf{\begin{tabular}[c]{@{}c@{}}48.98/\\ \underline{18.05}\end{tabular}} & \textbf{\begin{tabular}[c]{@{}c@{}}50.22/\\ 21.01\end{tabular}} & \textbf{\begin{tabular}[c]{@{}l@{}}48.32/\\ 20.06\end{tabular}} & \textbf{\begin{tabular}[c]{@{}r@{}}51.02/\\ 23.26\end{tabular}} & \textbf{\begin{tabular}[c]{@{}c@{}}\underline{47.90}/\\ 19.03\end{tabular}} & \textbf{\begin{tabular}[c]{@{}l@{}}51.91/\\ 24.53\end{tabular}} \\
\noalign{\vskip 0.9ex}
\multicolumn{2}{l}{YEM}    \begin{tabular}[c]{@{}l@{}}\textit{WER/}\\ \textit{CER}\end{tabular}                                                                       & \cellcolor{base}\begin{tabular}[c]{@{}l@{}}70.22/\\ 28.97\end{tabular} &\cellcolor{base} \begin{tabular}[c]{@{}l@{}}62.51/\\ 24.42\end{tabular} & \textbf{\begin{tabular}[c]{@{}r@{}}60.74/\\ 23.49\end{tabular}} & \textbf{\begin{tabular}[c]{@{}c@{}}60.35/\\ 22.97\end{tabular}} & \textbf{\begin{tabular}[c]{@{}c@{}}\underline{60.21}/\\ 23.28\end{tabular}} & \begin{tabular}[c]{@{}c@{}}64.82/\\ 26.51\end{tabular}          & \begin{tabular}[c]{@{}l@{}}62.60/\\ 25.47\end{tabular}          & \begin{tabular}[c]{@{}r@{}}63.32/\\ 27.30\end{tabular}          & \textbf{\begin{tabular}[c]{@{}c@{}}60.73/\\ \underline{23.01}\end{tabular}} & \begin{tabular}[c]{@{}l@{}}64.70/\\ 30.56\end{tabular}          \\
\noalign{\vskip 0.9ex}
\multicolumn{2}{l}{\textit{Avg Int}} \begin{tabular}[c]{@{}l@{}}\textit{WER/}\\ \textit{CER}\end{tabular}                                                        & \cellcolor{base}\begin{tabular}[c]{@{}l@{}}59.20/\\ 24.54\end{tabular} & \cellcolor{base}\begin{tabular}[c]{@{}l@{}}57.46/\\ 23.26\end{tabular} & \textbf{\begin{tabular}[c]{@{}r@{}}54.23/\\ 21.58\end{tabular}} & \textbf{\begin{tabular}[c]{@{}c@{}}53.39/\\ 19.97\end{tabular}} & \textbf{\begin{tabular}[c]{@{}c@{}}53.16/\\ \underline{19.76}\end{tabular}} & \textbf{\begin{tabular}[c]{@{}c@{}}54.98/\\ 22.42\end{tabular}} & \textbf{\begin{tabular}[c]{@{}l@{}}53.35/\\ 21.80\end{tabular}} & \textbf{\begin{tabular}[c]{@{}r@{}}54.51/\\ 23.48\end{tabular}} & \textbf{\begin{tabular}[c]{@{}c@{}}\underline{52.22}/\\ 20.01\end{tabular}} & \textbf{\begin{tabular}[c]{@{}l@{}}55.10/\\ 25.64\end{tabular}} \\
\noalign{\vskip 0.6ex} \cdashline{1-12}\noalign{\vskip 1.0ex}
\multicolumn{2}{l}{\textit{Avg All}}   \begin{tabular}[c]{@{}l@{}}\textit{WER/}\\ \textit{CER}\end{tabular}                                                           & \cellcolor{base}\begin{tabular}[c]{@{}l@{}}44.11/\\ 19.84\end{tabular} & \cellcolor{base}\begin{tabular}[c]{@{}l@{}}44.31/\\ 19.45\end{tabular} & \begin{tabular}[c]{@{}r@{}}48.60/\\ 25.61\end{tabular}          & \textbf{\begin{tabular}[c]{@{}c@{}}40.40/\\ 16.81\end{tabular}} & \textbf{\begin{tabular}[c]{@{}c@{}}\underline{40.13}/\\ \underline{16.48}\end{tabular}} & \textbf{\begin{tabular}[c]{@{}c@{}}43.21/\\ 18.85\end{tabular}} & \textbf{\begin{tabular}[c]{@{}l@{}}41.88/\\ 18.27\end{tabular}} & \textbf{\begin{tabular}[c]{@{}r@{}}42.67/\\ 19.36\end{tabular}} & \textbf{\begin{tabular}[c]{@{}c@{}}40.52/\\ 16.61\end{tabular}} & \textbf{\begin{tabular}[c]{@{}l@{}}41.11/\\ 19.32\end{tabular}} \\
\noalign{\vskip 0.6ex}
\multicolumn{2}{l}{\textit{Avg Dia}} \begin{tabular}[c]{@{}l@{}}\textit{WER/}\\ \textit{CER}\end{tabular}                                                        & \cellcolor{base}\begin{tabular}[c]{@{}l@{}}57.85/\\ 26.01\end{tabular} & \cellcolor{base}\begin{tabular}[c]{@{}l@{}}57.48/\\ 25.58\end{tabular} & \begin{tabular}[c]{@{}r@{}}58.66/\\ 29.10\end{tabular}          & \textbf{\begin{tabular}[c]{@{}c@{}}52.56/\\ 21.63\end{tabular}} & \textbf{\begin{tabular}[c]{@{}c@{}}52.40/\\ 21.36\end{tabular}} & \textbf{\begin{tabular}[c]{@{}c@{}}54.98/\\ 22.42\end{tabular}} & \textbf{\begin{tabular}[c]{@{}l@{}}53.35/\\ 21.80\end{tabular}} & \textbf{\begin{tabular}[c]{@{}r@{}}54.51/\\ 23.48\end{tabular}} & \textbf{\begin{tabular}[c]{@{}c@{}}\underline{52.22}/\\ \underline{20.01}\end{tabular}} & \textbf{\begin{tabular}[c]{@{}l@{}}55.10/\\ 25.64\end{tabular}} \\ \hline
\end{tabular}
}
\caption{WER ($\downarrow$) and CER ($\downarrow$) across various Arabic speech conditions using baseline and \textbf{context-aware Whisper decoding strategies}. Baseline models are \textbf{SM4T:} SeamlessM4T and \textbf{W-v3:} Whisper-large-v3. Our prompt-based methods ($\leftarrow$) inject contextual text into the decoder using \textbf{W←FPT:} first-pass transcriptions. $\leftarrow$\textbf{Rand:} randomly shuffling the prompt's words and $\leftarrow$\textbf{Rev:} reversing the prompt word's order. $\leftarrow$\textbf{SMms} and $\leftarrow$\textbf{SSea:} retrieving similar sentences based on MMS or SM4T, respectively. Prefix-based methods ($\rightarrow$) concatenate contextual (speech, text) pairs at the encoder/decoder inputs. \textbf{No-TTS:} retrieve a similar speech for the example. \textbf{TTS:} Speaker-conditioned synthesis.}
\label{tab:arabic-wer-cer}
\end{table*}

We evaluate context-aware decoding on ten Arabic conditions spanning MSA, accented MSA, and diverse dialectal speech. Experiments cover Common Voice 15.0 (CV)~\cite{ardila2019common}, FLEURS~\cite{conneau2023fleurs}, the MGB broadcast benchmarks~\cite{ali2016mgb,ali2017speech,ali2019b}, and five curated conversational dialect datasets (in-house data in~\newcite{waheed2024distill}). Dataset details are in Appendix~\ref{app:datasets}. All results use a shared text normalization pipeline (Appendix~\ref{app:preprocessing}); we also keep decoding hyperparameters fixed within each model family across all method variants. Because our methods target two ASR architectures, we report Whisper-based results and CTC-based results separately. We first analyze prompt- and prefix-based context integration for Whisper (Section~\ref{sec:context-aware-decoding}), summarized in Table~\ref{tab:arabic-wer-cer}, followed by proxy-guided n-best selection for non-autoregressive CTC models (Section~\ref{exp-ctc}). All numbers follow standardized text normalization (Appendix~\ref{app:preprocessing}).


\subsection{Context-Aware Decoding for Whisper}\label{sec:context-aware-decoding}
Table~\ref{tab:arabic-wer-cer} reports WER/CER for zero-shot Whisper decoding. We compare the standard Whisper-large-v3 baseline against prompt-based and prefix-based context-aware variants, and include SeamlessM4T (SM4T)~\cite{barrault2023seamlessm4t} as a strong standalone baseline. Unless stated otherwise, retrieval indices (text or speech-text) are fixed before evaluation and exclude evaluation transcripts to avoid leakage. 

\paragraph{Baselines.}
Consistent with prior work~\cite{waheed2024udistil}, SM4T outperforms Whisper on most dialectal conditions, while both models achieve substantially lower error rates on MSA than on dialectal Arabic, highlighting the persistent gap between standard and non-standard varieties.

\paragraph{MSA and accented MSA.}
On MSA and accented MSA, prompt-based integration can yield large gains only when prompt structure is carefully controlled. Direct prompting with first-pass transcriptions frequently triggers prompt continuation and hallucinations failures, particularly for incomplete or weakly constrained utterances. In such cases, Whisper may over-interpret the prompt as a coherent text to be continued rather than auxiliary context, producing empty outputs or boilerplate filler. In Arabic, recurring hallucinations include phrases such as \<اشتركوا في القناة> (``subscribe to the channel") or \<ترجمة نانسي قنقر> (``Translated by Nancy Kangar"), consistent with prior observations on Arabic Whisper failures~\cite{talafha2023n}.\footnote{Similar artifacts have been reported across languages (e.g., subtitle tags or generic markers), suggesting a borader weakness of autoregressive decoding rather than a language-specific phenomenon; see, e.g., \url{https://gist.github.com/riotbib/3b3c5f817b55b68801d14b8bdb02df09} and \url{https://medium.com/@lehandreassen/who-is-nicolai-winther-985409568201}.}

Prompt reordering strategies—random shuffling or reversing the prompt—substantially suppress hallucination behaviors by disrupting sequential coherence. Across MSA and accented MSA, reversed prompts yield the most reliable improvements, outperforming both standard Whisper and direct prompting while minimizing the abovementioned common hallucination patterns. To better understand this effect,  qualitative examples are provided in Appendix~\ref{app:hallucination}.



Retrieval-based prompting provides modest MSA gains when lexical alignment between the prompt and target utterance is strong, but degrades under domain mismatch (notably on FLEURS). Prefix-based methods are effective only when speaker characteristics are aligned: using retrieved audio without adaptation offers limited benefit, whereas speaker-consistent synthesized prefixes yield consistent improvements on MSA. Overall, prompt reversal emerges as the most robust adaptation strategy for accented MSA, while speaker-consistent prefixing is most effective on MSA.

\paragraph{External dialectal datasets (MGB-3/5).} Whisper performs poorly on external dialectal benchmarks, with baseline WERs of 35.90\% on MGB-3 (Egyptian) and 79.16\% on MGB-5 (Moroccan), reflecting severe dialect and domain mismatch. First-pass direct prompting exacerbates this issue on MGB-3, increasing WER to 63.47\%, as Whisper often treats the prompt as ground truth and produces empty or boilerplate outputs. On MGB-5, direct prompting yields only marginal gains (76.04\%).

Prompt reordering substantially mitigates these failures. Random shuffling reduces WER to 31.62\% on MGB-3 and 69.37\% on MGB-5, while reversing achieves the best result on MGB-3 (31.14\%) and comparable performance on MGB-5 (69.89\%). These gains indicate that disrupting prompt syntax discourages prompt-continuation artifacts in highly mismatched dialectal settings. 

Retrieval-based prompting yields limited benefits. MMS-based retrieval slightly degrades MGB-3 performance and provides only marginal gains on MGB-5, while SM4T-based retrieval narrows this gap but remains consistently weaker than prompt reordering. Prefix-based methods are more stable on dialects: prefixing with raw retrieved audio provides modest gains but remains sensitive to speaker mismatch. In contrast, speaker-consistent, synthesized prefix improves robustness, reducing WER to 34.22\% on MGB-3 and 73.70\% on MGB-5. The strongest prefix-based result is obtained by self-prefixing, which reduces MGB-5 WER to 68.21\% ($\approx$14\% relative improvement). Averaged across both datasets, prompt reordering remains the most effective strategy.

\paragraph{Casa-\textit{Dialects}.} Across five dialectal test sets (Algerian, Jordanian, Palestinian, Emirati, and Yemeni), Whisper performs substantially worse than on MSA, with an average WER/CER of 57.46/23.26 compared to 15.79/6.35 on MSA. Error rates vary widely across dialects, with Algerian being the most challenging (78.60/37.81) and Jordanian the least (40.79/13.55), reflecting a combination of linguistic divergence and domain mismatch.

Prompting with SM4T first-pass hypothesis reduces the average error to 54.23/21.58, with gains concentrated in dialects closer to MSA, such as Jordanian (37.34/12.12), Palestinian (46.12/14.98), and Emirati (49.10/18.13). Prompt-reordering further improves robustness: shuffling is particularly effective for Algerian, reducing WER to 73.07\%, while reversing yields the best overall average (53.16/19.76), corresponding to a relative WER reduction of nearly 7.5\% over Whisper's baseline. 

Retrieval-based prompting provides limited benefits. Retrieval using SM4T transcripts yields a modest improvement (53.35/21.80), whereas MMS-based retrieval is less effective, again indicating that lexical overlap with the greatest variety (i.e., MSA) plays a larger role than semantic similarity alone. Prefix-based methods show mixed behavior. Prefixing with retrieved audio degrades performance (54.51/23.48), often due to speaker mismatch. In contrast, speaker-consistent, synthesized prefixes substantially improve stability and yield the best overall performance across Casa-\textit{Dialects}, averaging (52.22/20.01, $\approx$9\% relative WER reduction), including improvements on difficult dialects such as Algerian (74.53/31.28). Self-prefixing achieves the largest gains on Algerian, yielding a $\approx$4.5\% WER reduction over retrieval-based prefixing.

\subsection{Proxy-Guided $n$-Best Selection for CTC ASR}
\label{exp-ctc}

\begin{figure}[t]
\centering
\includegraphics[width=\linewidth]{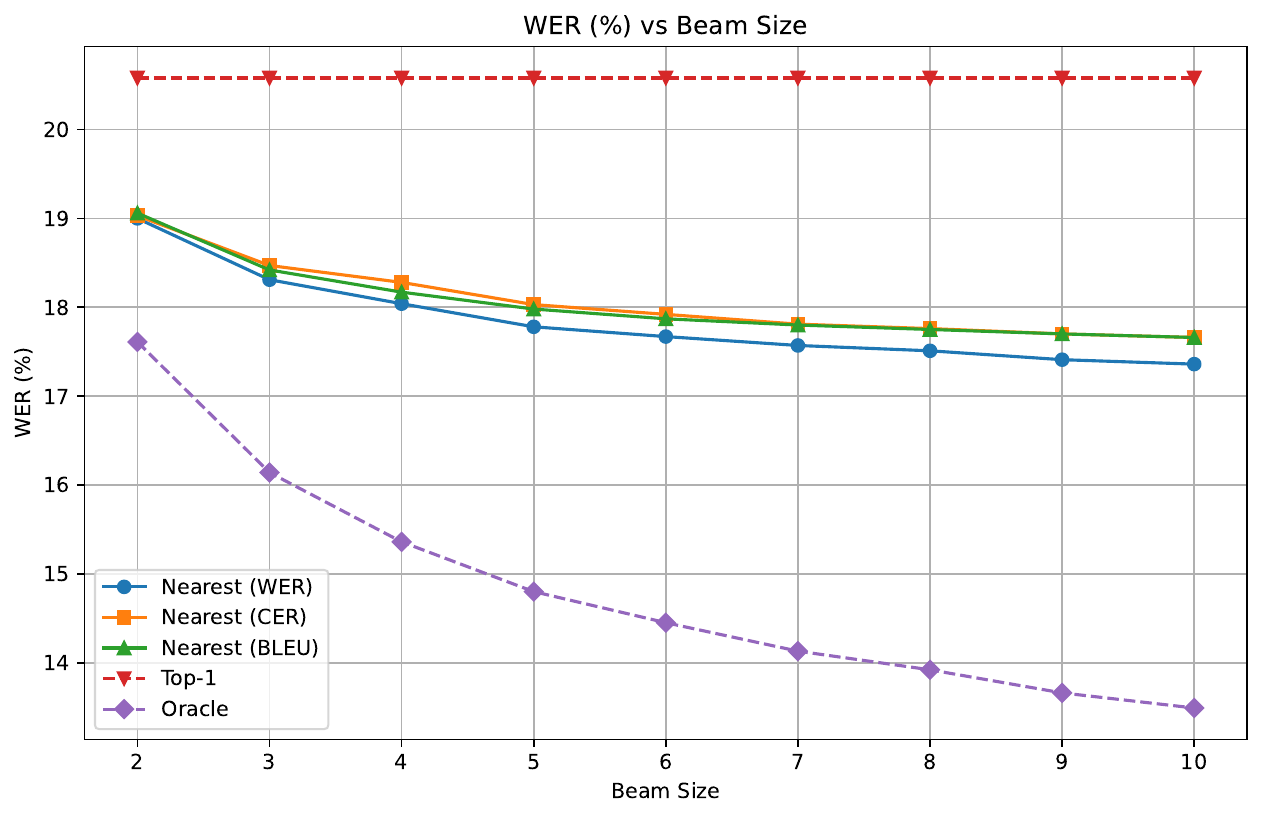}
\caption{WER of Omnilingual ASR (CTC) on MSA as a function of beam size. Proxy-guided $n$-best selection (\textbf{Nearest}) consistently outperforms top-1 decoding and recovers a large fraction of the oracle $n$-best gains.}
\label{fig:xlsr_reranking}
\end{figure}
We next evaluate whether external proxy transcriptions can improve decoding for non-promptable, non-autoregressive (CTC) ASR models. Experiments use Omnilingual  ASR (\texttt{omniASR\_CTC\_1B})\footnote{\url{https://github.com/facebookresearch/omnilingual-asr}} with a CTC objective on Common Voice MSA as a controlled testbed with stable decoding behavior.

For each utterance, the CTC model produces an $n$-best list using beam search with beam sizes $B \in \{2,\ldots,10\}$. In parallel, auxiliary ASR systems generate first-pass transcriptions that serve as external proxies. For a given beam size, we select candidates whose text is \textit{closest to the proxy} under a distance metric $d \in \{\text{WER}, \text{CER}, 1{-}\text{BLEU}\}$, which we refer to as \textbf{Nearest} selection. As an upper bound, we also report an \textbf{Oracle} that selects the hypothesis with the lowest WER to the reference.

Figure~\ref{fig:xlsr_reranking} plots WER as a function of beam size for top-1 decoding, proxy-guided Nearest selection, and the Oracle. Top-1 decoding remains fixed at 20.58\% WER across all beam sizes, indicating that increasing the beam primarily expands lower-ranked alternatives while leaving the highest-scoring hypothesis unchanged for most utterances. In contrast, proxy-guided selection consistently reduces error rates: at beam size 2, Nearest lowers WER to 19.00\%, and at beam size 10 it reaches 17.36\%, corresponding to a 15.7\% relative reduction over top-1 decoding.

The Oracle curve highlights the remaining headroom in the $n$-best list. At beam size 10, the Oracle achieves 13.49\% WER, indicating that proxy-guided selection recovers 70\% of the oracle improvement. Results are consistent across WER-, CER-, and BLEU-based distances, with WER-based selection performing marginally better, suggesting that proxy transcriptions provide a robust guidance signal largely insensitive to the specific similarity metric.

\section{Discussion}
This section provides additional analysis to clarify (i) the effect of speaker-conditioned TTS on prefix-based context, (ii) design choices for text retrieval in prompt-based conditioning, and (iii) when multi-proxy guidance helps proxy-guided $n$-best selection.

\subsection{Impact of Speaker-Conditioned TTS on Prefix-Based Decoding}
\label{sec:tts-impact}

Prefix-based context integration can use either a \emph{real} contextual waveform $\mathbf{x}_{\mathrm{ctx}}$ (retrieved from a speech--text collection) or a \emph{synthetic} contextual waveform $\tilde{\mathbf{x}}_{\mathrm{ctx}}$ generated by speaker-conditioned TTS from the same contextual transcript $\mathbf{p}_{\mathrm{ctx}}$. A natural concern is that TTS artifacts might degrade Whisper's recognition accuracy by introducing acoustic mismatch or unnatural prosody.

To assess this effect, we compare prefix-based decoding using real versus TTS-generated contextual audio on three datasets (CV15, MGB-3, and FLEURS). Figure~\ref{fig:tts-chart} (Appendix) shows that replacing real contextual speech with synthesized speech increases WER by a modest amount: $+6.79$ WER points on CV15, $+4.98$ on MGB-3, and $+1.05$ on FLEURS. Averaged across datasets, the degradation remains within $4.27$ WER points and $3.41$ CER points. Overall, these results indicate that, when paired with speaker-conditioned synthesis, our TTS pipeline preserves sufficient acoustic fidelity for contextual prefixing and does not materially change Whisper's decoding behavior. This supports using synthesized context as a practical alternative to relying on matched, labeled contextual speech exemplars.

\subsection{Retrieval Representations: Why Character-Level TF--IDF Works Best}
\label{exp-similarity}

For prompt retrieval, we evaluate four representations for measuring similarity between a first-pass hypothesis and candidates in a fixed text index: (i) character-level TF--IDF, (ii) dense text embeddings, (iii) speech-derived embeddings, and (iv) speaker embeddings. We compare these choices using downstream ASR performance after injecting the retrieved sentence as a decoder prompt.

As shown in Table~\ref{tab:feature-extractor-results} (Appendix; see also Appendix~\ref{app:which-mode}), character-level TF-IDF is the most effective retrieval representation, reducing WER from 22.84\% (standard Whisper decoding) to 17.89\%. In contrast, dense text embeddings reach 20.04\% WER, while speech- and speaker-based embeddings are less effective (24.78\% and 27.16\% WER, respectively). TF-IDF’s strength lies in \emph{lexical overlap}, robustness to orthographic variation, and low cost, making it effective for dialectal Arabic and noisy hypotheses. In addition, TF-IDF operates purely in the text domain and is computationally inexpensive, motivating its use as the default prompt-retrieval method in our experiments.

\subsection{Multi-Proxy Interpolation for Proxy-Guided $n$-Best Selection}
\label{sec:multi-proxy}

\paragraph{Quantitative trends.}
We extend proxy-guided $n$-best selection by interpolating guidance signals from multiple auxiliary ASR systems. Given two proxy transcriptions $p_1$ and $p_2$, we score each candidate hypothesis $h$ using a weighted sum of distances:
$d(h) = \alpha \, d(h, p_1) + (1-\alpha)\, d(h, p_2)$, where $d(\cdot,\cdot)$ denotes a hypothesis-to-hypothesis \emph{normalized word-level edit distance} (WED; i.e., WER-style edit distance computed between two strings) and $\alpha \in [0,1]$ controls the contribution of each proxy. We then select the hypothesis with the lowest interpolated score. Figure~\ref{fig:div-chart} (Appendix) summarizes interpolation behavior on MSA and dialectal test sets. Single-proxy selection already improves over top-1 CTC decoding across all datasets, and multi-proxy interpolation yields additional gains that are modest but stable. For example, on CV (MSA), top-1 decoding yields 20.58\% WER, single-proxy selection reduces WER to 17.36\%, and interpolation further reduces WER to 17.34\%. This additional improvement is small relative to the remaining gap to the oracle (13.49\%), but consistent across a range of interpolation weights. Similar trends hold for dialectal Arabic. See Appendix~\ref{app:interpolation} fo details and for a discussion illustrated with examples of why interpolation helps. 

\section{Conclusion}

We studied \emph{context-aware decoding} as a lightweight, training-free test-time adaptation paradigm for zero-shot Arabic ASR. Across MSA, accented MSA, and diverse dialectal conditions, our methods improve recognition without parameter updates or architecture changes by injecting contextual side information at inference time. For promptable encoder--decoder ASR (Whisper), structured prompt- and prefix-based integration can substantially reduce error rates: prompt reordering mitigates prompt-continuation hallucinations, and speaker-consistent prefixing improves stability when contextual exemplars differ in speaker characteristics. Beyond promptable architectures, we showed that contextual guidance also benefits non-autoregressive CTC ASR via proxy-guided $n$-best selection, which consistently outperforms top-1 decoding and recovers a substantial portion of the available oracle $n$-best improvement. Using multiple auxiliary proxies provides small but reliable additional gains in our interpolation sweeps, suggesting that different ASR systems offer complementary error signals. Future work will explore: (i) prompt-aware fine-tuning to reduce hallucinations; (ii) retrieval and prompting for code-switching and domain mismatch; and (iii) stronger proxy reranking (e.g., LLM-based) and extending contextualization to emerging promptable ASR backbones.

\section*{Limitations}
Despite the consistent gains from context-aware decoding, several limitations remain. These limitations were most salient in settings with strong mismatch (e.g., retrieval under domain shift on FLEURS and highly mismatched dialectal broadcast speech) and may affect scalability, latency, and generalization in real deployments.

\begin{enumerate}
\item \textbf{Computation and latency overhead.}
Our methods add per-utterance steps beyond standard decoding, including (i) generating first-pass hypotheses with an auxiliary ASR, (ii) feature extraction and nearest-neighbor search for retrieval, and (iii) optional speaker-conditioned TTS for constructing contextual prefixes. These components increase compute and can introduce additional latency, which may be undesirable for streaming, real-time, or edge scenarios. Similar overhead has been reported for retrieval-augmented or $k$NN-augmented Whisper decoding~\cite{wang2024can,nachesa2024knn,shen2025retrieval}.

\item \textbf{Sensitivity to auxiliary component quality.}
The effectiveness of contextual prompting, prefixing, and proxy-guided selection depends on the quality of auxiliary signals. Noisy proxy hypotheses can mislead prompting or reranking (e.g., exacerbating prompt-continuation artifacts under direct prompting), and TTS artifacts (pronunciation errors, prosody mismatch, or limited dialectal coverage) can reduce the usefulness of synthesized prefixes. While our analyses suggest that speaker-conditioned synthesis is reasonably robust on the evaluated sets, stronger mismatch or lower-quality TTS may degrade performance.

\item \textbf{Limited effective prompt budget.}
Promptable ASR models impose practical limits on how much context can be supplied at inference time. In common Whisper implementations, only a bounded number of prompt tokens are effectively used during decoding (often on the order of a few hundred tokens), which constrains the benefit of long contextual inputs.\footnote{\url{https://platform.openai.com/docs/guides/speech-to-text}} Some toolchains expose larger context windows, but a nontrivial portion is reserved for task tokens and decoding context.\footnote{\url{https://github.com/huggingface/transformers/issues/27445}} This limitation motivates future work on selective prompting and compact, structured contextual cues.

\item \textbf{Dialect and domain coverage.}
Although we evaluate across multiple datasets and dialects, Arabic remains underrepresented in public ASR resources. Several dialects (e.g., Sudanese, Mauritanian, Iraqi) are not covered in our experiments, and existing benchmarks may exhibit domain, genre, or demographic skew. Consequently, the magnitude of gains may vary when transferring to unseen dialects, spontaneous conversational speech, or heavy code-switching.

\item \textbf{Retrieval dependence and scalability.}
Retrieval quality depends on corpus composition, normalization, and the similarity function. Lexical retrieval (e.g., character-level TF--IDF) can be sensitive to tokenization and spelling variation, while semantic or acoustic retrieval may overemphasize particular domains or speaker traits. Moreover, scaling to larger corpora can improve recall but increases indexing and search-time costs, which may further impact deployment constraints.

\item \textbf{Incomplete exploration of prompting and reranking strategies.}
Our study focuses on TF--IDF-based retrieval and simple prompt reordering (reverse/shuffle), as well as distance-based proxy reranking. Many alternatives remain to be explored, including confidence-weighted or selective prompting, structured prompts that explicitly encode dialect or topic, learned rerankers, and LLM-based context generation that produces more constrained, dialect-aware cues~\cite{suh2024}.
\end{enumerate}

\bibliography{acl_latex}

\appendix

\definecolor{Green}{HTML}{32CD32}
\definecolor{Plum}{HTML}{DDA0DD}
\definecolor{LimeGreen}{HTML}{FEE685}

\section{Appendix}

\begin{figure*}[t]
\centering
\includegraphics[width=1\textwidth]{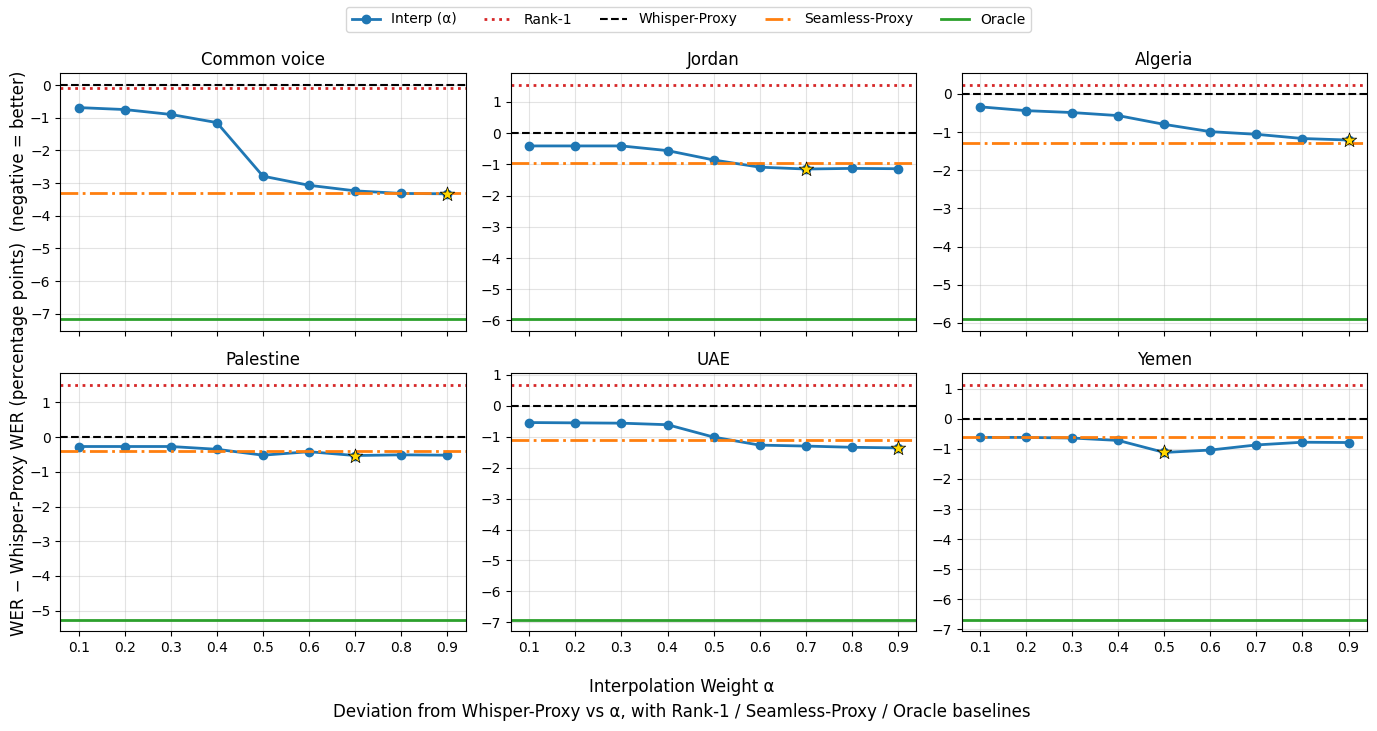}
\caption{Deviation in WER (percentage points) relative to single-proxy Whisper selection as a function of interpolation weight 
$\alpha$, across MSA (Common Voice) and dialectal Arabic test sets. Negative values indicate improvement. Multi-proxy interpolation consistently improves over single-proxy selection and remains stable across a wide range of 
$\alpha$, approaching the oracle bound without requiring careful tuning.}
\label{fig:div-chart}
\end{figure*}

\begin{table*}[t!]
\centering
\footnotesize
\setlength{\tabcolsep}{4pt}
\renewcommand{\arraystretch}{1.15}
\resizebox{\textwidth}{!}{%
\begin{tabular}{lllll}
\toprule
\textbf{Reference} & \textbf{TF-IDF} & \textbf{Text Embedding} & \textbf{Speech Embeddings} & \textbf{Speaker Embeddings} \\
\midrule
(\<من الممكن انها لن تاتي غدا>) & (\<من الممكن انها ستاتي>) & (\<لم لن تكون هنا غدا>) & (\<ربما من الافضل ان تاتي معنا>) & (\<وداعيا الى الله باذنه وسراجا منيرا>) \\
(\<لقد قابلته>) & (\<قابلته يوما ما>) & (\<قابلته يوما ما>) & (\<اغنى خلقه بالمال>) & (\<ساخذه>) \\
(\<انك تكبر المشكلة>) & (\<انا لست المشكلة>) & (\<هو في مشكلة>) & (\<وتشير باليد>) & (\<واجتباه لنبوته>) \\
(\<درس كل يوم لمدة ساعة ونصف>) & (\<وجع ساعة ولا كل ساعة>) & (\<لماذا تدرس كل يوم>) & (\<نعم سيكفي الحليب حتى يوم الجمعة>) & (\<عند سدرة المنتهى>) \\
(\<ذهبت الى هناك ايضا>) & (\<انا ايضا>) & (\<اذهب الى هناك الان>) & (\<يجب ان تذهب غربا>) & (\<ترك لي توم رسالة>) \\
(\<اتحسب سوء الظن يجرح في فكري>) & (\<فاما سوء الظن بها فقد اختلف الناس فيه>) & (\<كان سوء الظن بها يعمي عن محاسنها>) & (\<انقذ الطفل بالمجازفة بحياته>) & (\<عن ابي محمد الحسن بن علي بن ابي طالب>) \\
\bottomrule
\end{tabular}
}
\caption{Examples of top retrieved sentences using different extractors. TF-IDF consistently preserves surface forms, while dense and acoustic features tend to retrieve semantically related but lexically or contextually divergent content. Sample size=1000}
\label{tab:which-mode}
\end{table*}

\subsection{Datasets}
\label{app:datasets}
\paragraph{Common Voice 15.0 (CV15).} A crowd-sourced dataset of read Arabic speech~\cite{ardila2019common}. Utterances written in \emph{MSA}, the formal variety used widely across the Arab world in news broadcasts, education, and official contexts.

\paragraph{MGB-2/3/5.} This collection comes from the Arabic Multi-Genre Broadcast (MGB) challenges~\cite{ali2016mgb,ali2017speech,ali2019b}, which feature speech from real-world broadcast content. MGB-2 (around 1,200 hours) contains \emph{MSA} with other dialects mixed in. MGB-3 ($\approx$6 hours) focuses on \emph{Egyptian dialect}, while MGB-5 ($\approx$6 hours) focuses on \emph{Moroccan Arabic}. We present MGB-3 and MGB-5 as \emph{external dialectal data}. We manually validated MGB samples and found errors like omissions, mismatches, and typos.

\paragraph{FLEURS.} The Arabic portion of FLEURS~\cite{conneau2023fleurs}. Features read speech sourced from news and web content. The speech is in \emph{MSA} but spoken with an Egyptian accent, as known as \emph{accented MSA}~\cite{waheed2024,talafha2023n}.

\paragraph{Casa-\textit{Dialects}.} 
In-House dataset presented in~\newcite{waheed2024distill} covering five underrepresented Arabic dialects: Algerian (ALG), Jordanian (JOR), Palestinian (PAL), Emirati (UAE), and Yemeni (YEM), representing four major regions (North African, Levantine, Gulf, and Yemeni). Native speakers annotated segments from YouTube-sourced local TV series. The resulting corpus contains 10,567 utterances and 121,293 words, totaling over 13 hours of speech. Detailed statistics are provided in ~\newcite{waheed2024distill}.


\subsection{Preprocessing} 
\label{app:preprocessing}
Some of the datasets include inconsistencies in formatting and script usage. For instance, certain utterances are fully marked with diacritics while others, sometimes from the same source, lack them entirely. To ensure consistency across all inputs, we apply a standard preprocessing pipeline inspired by~\newcite{talafha2023n}. Specifically, we remove all punctuation except the \% and @ symbols, strip diacritics, Hamzas, and Maddas, and convert Eastern Arabic numerals to their Western equivalents (e.g., \<٩٢> to 29). Additionally, since our focus is not on code-switching, we exclude any Latin-script tokens from the data.

\subsection{Model Settings}
\label{app:settings}
All experiments were conducted using the \textit{transformers} and \textit{datasets} libraries from HuggingFace. All audio segments were resampled to a sampling rate of 16kHz. Evaluations were performed on a single computing node equipped with 8 A10 GPUs (24GB each). For ASR systems, we employed: \textit{Whisper:} \texttt{whisper-large-v3}\footnote{\url{https://huggingface.co/openai/whisper-large-v3}} (1.55B parameters), \textit{SeamlessM4T:} \texttt{seamless-m4t-v2-large}\footnote{\url{https://huggingface.co/facebook/seamless-m4t-v2-large}} (2.3B parameters), \textit{MMS:} \texttt{mms-1b-all}\footnote{\url{https://huggingface.co/facebook/mms-1b-all}} (1B parameters), and \textit{Omnilingual}: \texttt{omniASR\_CTC\_1B	}\footnote{\url{https://github.com/facebookresearch/omnilingual-asr}} (1B parameters)\\
\\
\noindent
For the retrieval-based components, we adopted the following extractors: \textbf{TF-IDF:} Character-level n-gram features using \texttt{analyzer="char\_wb"} and \texttt{ngram\_range=(3, 5)}. \textbf{Sentence Embeddings:} We used an off-the-shelf Arabic sentence encoder, \footnote{\url{https://huggingface.co/Omartificial-Intelligence-Space/Arabic-mpnet-base-all-nli-triplet}}, \textbf{Speech Embeddings:} Extracted from the final hidden states of the \texttt{whisper-large-v3} encoder. \textbf{Speaker Embeddings:} Derived from speaker verification with ECAPA-TDNN embeddings\footnote{\url{https://huggingface.co/speechbrain/spkrec-ecapa-voxceleb}} trained on Voxceleb dataset~\cite{desplanques2020ecapa}. All models were used with their default hyperparameter settings unless otherwise specified.

\subsection{Effect of Reversed Prompting on Hallucination and Output Fidelity}
Table~\ref{tab:qual-examples} presents manually selected examples illustrating the impact of reversed prompting on transcription quality. In each case, we compare the output of Whisper when conditioned on a standard SM4T-based prompt versus a reversed version of the same prompt. The examples highlight failure modes such as hallucinated phrases or overly short outputs in the standard prompt condition. Reversed prompting consistently recovers content that is more faithful to the reference transcription, with substantially lower WER.

\begin{table}[h!]
\centering
\renewcommand{\arraystretch}{0.5}   
\resizebox{1.0\linewidth}{!}{
\begin{tabular}{lp{8cm}}
\toprule
\cellcolor{Green!30}\textbf{\textit{Example 1}}  & \cellcolor{Green!30} \RaggedLeft{\rule{0pt}{1.2em}\textbf{}} \\[0.2cm]  \toprule 
\cellcolor{LimeGreen!30}\textbf{Reference} & \cellcolor{LimeGreen!30} \RaggedRight{\begin{arabtext} \small  
وايضا اعطى للعملية برمتها نوع من الـ
\end{arabtext}} \\  

\cellcolor{LimeGreen!30}\textbf{Whisper+prompt} & \cellcolor{LimeGreen!30} 
\textcolor{red}{
\RaggedRight{\begin{arabtext} \small  
مستشوفيات كثيرة في
\end{arabtext}}} \\ 

\cellcolor{LimeGreen!30}\textbf{Whisper+Rev} & \cellcolor{LimeGreen!30} \RaggedRight{\begin{arabtext} \small  
وايضا اعطى للعملية برمتها نوع من
\end{arabtext}} \\

\cellcolor{LimeGreen!30}\textbf{WER (prompt)} & \cellcolor{LimeGreen!30} \RaggedRight{\begin{arabtext} 
$1.00$
\end{arabtext}} \\

\cellcolor{LimeGreen!30}\textbf{WER (Rev)} & \cellcolor{LimeGreen!30} \RaggedRight{\begin{arabtext} 
$0.14$
\end{arabtext}} \\ 
\\[0.1cm]
\toprule
\cellcolor{Green!30}\textbf{\textit{Example 2}}  & \cellcolor{Green!30} \RaggedLeft{\rule{0pt}{1.2em}\textbf{}} \\[0.2cm]  \toprule 
\cellcolor{LimeGreen!30}\textbf{Reference} & \cellcolor{LimeGreen!30} \RaggedRight{\begin{arabtext} \small  
ان الـ الخطاب السياسي مثلا اصبح مقيدا من طرف القضاء\end{arabtext}} \\ 

\cellcolor{LimeGreen!30}\textbf{Whisper+prompt} & \cellcolor{LimeGreen!30} \textcolor{red}{\RaggedRight{\begin{arabtext} \small  
بل\end{arabtext}}} \\ 

\cellcolor{LimeGreen!30}\textbf{Whisper+Rev} & \cellcolor{LimeGreen!30} \RaggedRight{\begin{arabtext} \small  
ان الخطاب السياسي مثلا اصبح مقيدا من طرف القضاء\end{arabtext}} \\

\cellcolor{LimeGreen!30}\textbf{WER (prompt)} & \cellcolor{LimeGreen!30} \RaggedRight{\begin{arabtext} 
$1.00$
\end{arabtext}} \\

\cellcolor{LimeGreen!30}\textbf{WER (Rev)} & \cellcolor{LimeGreen!30} \RaggedRight{\begin{arabtext} 
$0.10$
\end{arabtext}} \\ 

\\[0.1cm]
\toprule
\cellcolor{Green!30}\textbf{\textit{Example 3}}  & \cellcolor{Green!30} \RaggedLeft{\rule{0pt}{1.2em}\textbf{}} \\[0.2cm]  \toprule 
\cellcolor{LimeGreen!30}\textbf{Reference} & \cellcolor{LimeGreen!30} \RaggedRight{\begin{arabtext} \small  
عبر دستور 2011 وعبر العمل الحكومي بهذه الطريقة اي
\end{arabtext}} \\ 

\cellcolor{LimeGreen!30}\textbf{Whisper+prompt} & \cellcolor{LimeGreen!30} \textcolor{red}{\RaggedRight{\begin{arabtext} \small  
اشتركوا في القناة\end{arabtext}}} \\ 

\cellcolor{LimeGreen!30}\textbf{Whisper+Rev} & \cellcolor{LimeGreen!30} \RaggedRight{\begin{arabtext} \small  
عبر دستور 2011 وعبر العمل الحكومي بهذه الطريقة\end{arabtext}} \\

\cellcolor{LimeGreen!30}\textbf{WER (prompt)} & \cellcolor{LimeGreen!30} \RaggedRight{\begin{arabtext} 
$1.00$
\end{arabtext}} \\ 

\cellcolor{LimeGreen!30}\textbf{WER (Rev)} & \cellcolor{LimeGreen!30} \RaggedRight{\begin{arabtext} 
$0.11$
\end{arabtext}} \\

\bottomrule
\end{tabular}
}
\caption{Manually selected examples showing how reversed prompting mitigates hallucinations and improves WER. }
\label{tab:qual-examples}
\end{table}

\begin{table}[h!]
\small
\centering
\begin{tabular}{lll}
\toprule
Mode & \hspace{1cm} & WER/CER \\
\midrule
\textit{Vanilla}         & \hspace{2cm} & 22.84/9.65 \\
\cdashline{1-3}
\noalign{\vskip 0.6ex}
\textbf{TFIDF}  & \hspace{2cm} & \textbf{17.89/7.96} \\
Text Embedding  & \hspace{2cm} & 20.04/7.83 \\
Speech          & \hspace{2cm} & 24.78/11.08 \\
Speaker         & \hspace{2cm} & 27.16/13.26 \\
\bottomrule
\end{tabular}
\caption{WER/CER using different feature extractors for text retrieval on CV15 (sample size = 1000).}
\label{tab:feature-extractor-results}
\end{table}

\subsection{Qualitative Analysis of Retrieval Modes}
\label{app:which-mode}

\noindent
We manually analyzed 1,000 samples from the CV15 dev set to better understand the behavior of different retrieval extractors. Table~\ref{tab:which-mode} presents six representative query sentences along with the top matches returned by each method. TF-IDF consistently retrieved sentences with higher token-level overlap with the reference, resulting in more aligned surface-level matches. In contrast, dense text embeddings often returned semantically related but lexically divergent paraphrases, while speech and speaker embeddings frequently retrieved contextually unrelated content due to acoustic or speaker similarity. It is important to note that retrieval comparisons are based on the first-pass transcription, which serves as the input to the retrieval system. These qualitative observations align with our quantitative results, where TF-IDF achieved the lowest WER and CER on CV15 (17.89 / 7.96; $n{=}1000$; see Table~\ref{tab:feature-extractor-results}). For example, when querying with the sentence (\<من الممكن انها لن تاتي غدا>),  TF-IDF retrieves the closely related (\<من الممكن انها ستاتي>),  maintaining both structural and lexical overlap. In contrast, the text embedding method returns (\<لم لن تكون هنا غدا>),  semantically related but lexically distinct, while the speech-based method yields the more generic (\<ربما من الافضل ان تاتي معنا>),  and the speaker-based method retrieves (\<وداعيا الى الله باذنه وسراجا منيرا>),  which shares little contextual relevance. A similar pattern is seen for the query (\<انك تكبر المشكلة>),  where TF-IDF returns the precise phrase (\<انا لست المشكلة>),  while speech and speaker retrievals yield vague or acoustically aligned but semantically distant matches.

\subsection{Qualitative Analysis of Hallucination Suppression}
\label{app:hallucination}. We manually inspected 30 development samples where sentence-level WER dropped from $\geq$1 to $0$ after prompt reversal. Hallucinations occur most often for incomplete utterances, background music, voice-over segments, or multi-speaker dialogue. For example, the truncated utterance \<وايضا اعطى للعملية برمتها نوع من الـ> (``It also gave the whole process a kind of") lacks semantic closure and usually triggers hallucinations under direct prompting. Prompt reordering, particularly reversal, suppresses these failures by reducing prompt-continuation tendencies (Table~\ref{tab:qual-examples}).

\subsection{TTS Efficiency}
\begin{figure}[t!]
\centering
\includegraphics[width=0.5\textwidth]{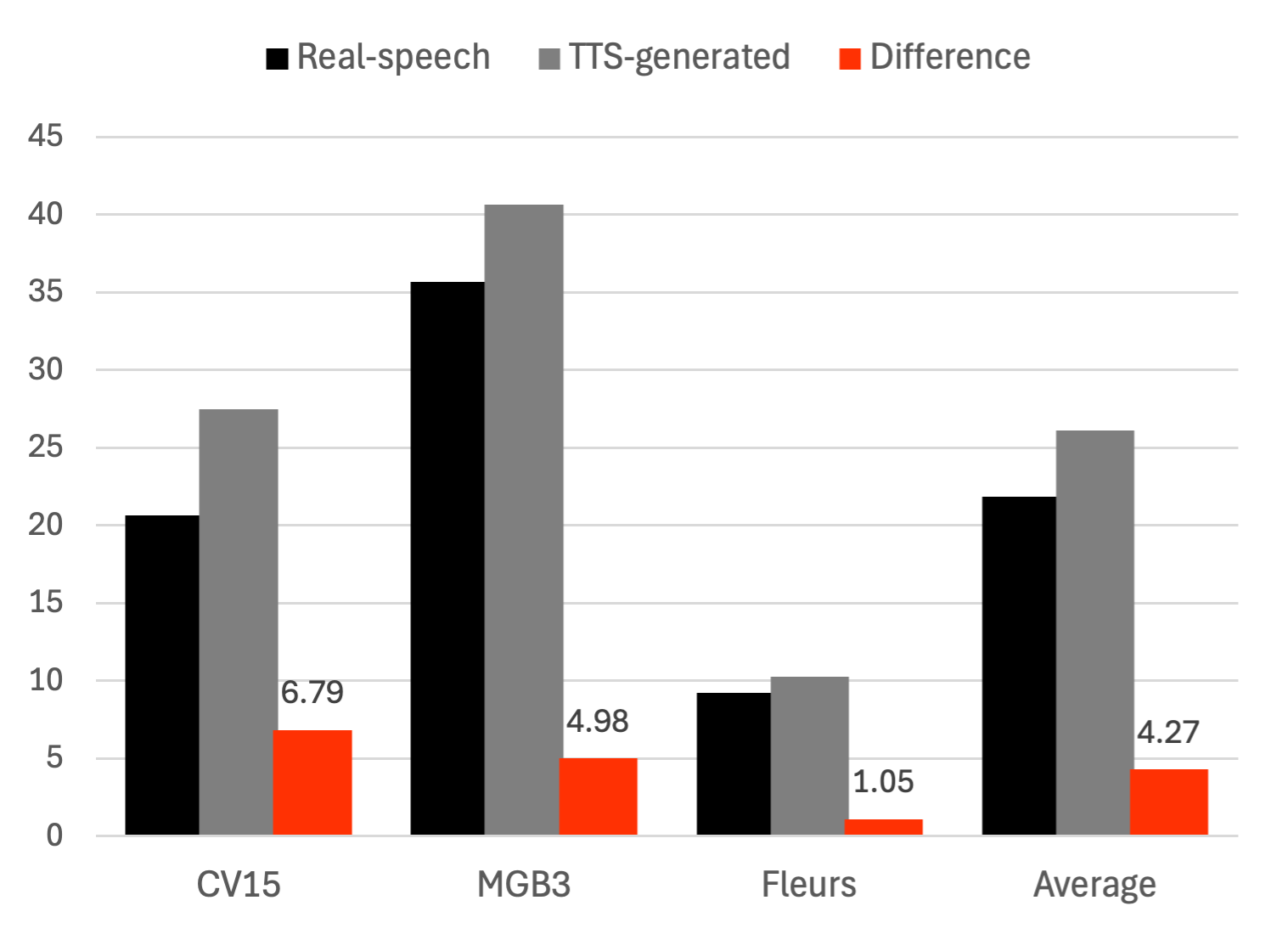}
\caption{Comparison between the performance of Whisper on real speech vs. TTS-generated speech across different language settings (sample size=1000).}
\label{fig:tts-chart}
\end{figure}

We measure the performance of the TTS model by transcribing its synthetic output and comparing it with real speech under three conditions (i.e., MSA, Dialect, Accented MSA). See Figure~\ref{fig:tts-chart}.

\subsection{Impact of Proxy Interpolation on CTC Decoding}
\label{app:interpolation}

In dialectal Arabic, interpolation improves over the stronger single-proxy baseline by approximately 0.5-2.0 WER points depending on the dialect. Performance is broadly stable across $\alpha$, with best values often occurring in the mid-to-high range (approximately $\alpha \in [0.5,0.9]$ in our sweeps). For Algerian, where the CTC baseline WER is very high (80.15\%), improvements are noise-limited; nevertheless, both single- and multi-proxy selection consistently outperform top-1 decoding.
\paragraph{Why interpolation helps.}
Table~\ref{tab:alpha-examples} presents representative cases in which interpolation selects hypotheses that are closer to the reference than both top-1 decoding and single-proxy selection. Across MSA and dialects (Jordanian, Palestinian, Emirati, Yemeni, Algerian), we observe that top-1 and single-proxy choices can share error modes, while interpolation is able to favor lower-ranked candidates that avoid these shared substitutions. For instance, in the MSA example, both top-1 decoding and Proxy~1 introduce an orthographic substitution in the main verb ((\<تعبت> → \<طعبت>)), while Proxy~2 corrects the verb but substitutes \<الرقود> with \<الركود>, altering meaning. Interpolation resolves this trade-off by selecting a lower-ranked beam (rank~3) that preserves both the correct verb form and the intended meaning. Similar behavior appears across dialects: in Jordanian speech, top-1 and proxy-based selections may alter noun inflection (\<شغلة> → \<شغلي>), whereas interpolation selects a higher-rank beam (rank~8) that restores the correct form. In Palestinian examples, phonetically similar but incorrect substitutions (e.g., \<أقوال> → \<أقواي>) are avoided by selecting lower-probability candidates (rank~10). In Emirati and Yemeni speech, top-1 and proxy-based outputs can introduce consonant or verb-prefix errors, while interpolation favors mid-ranked hypotheses (ranks~5--8) that recover grammatical verb forms. For Algerian, interpolation can likewise correct dialect-specific distortions by selecting a lower-ranked beam (rank~3). Overall, these examples suggest that combining complementary proxy signals can steer selection toward candidates that are more linguistically coherent than those favored by any single proxy.

\begin{table}[b!]
\centering
\renewcommand{\arraystretch}{0.1}   
\resizebox{1.0\linewidth}{!}{
\begin{tabular}{lp{11cm}}
\toprule 
\cellcolor{Green!30}
\textbf{\textit{Variety}}  & \cellcolor{Green!30} \RaggedLeft{\rule{0pt}{1.2em}\textbf{Jordan}} \\[0.2cm]  \toprule 
\cellcolor{LimeGreen!30} \textbf{Reference} & \cellcolor{LimeGreen!30} \RaggedRight{\begin{arabtext} \small  
أنا صار لي ساعة بحكي، وأنت بتقول لي بدي أفكر، هاي شغلة ما بدهاش تفكير
\end{arabtext}} \\  

\cellcolor{LimeGreen!30} \textbf{Rank-1} & \cellcolor{LimeGreen!30} \RaggedRight{\begin{arabtext} \small  
أنا صار لي ساعة بحكي وأنت بتقول لي بدي أفكر هاي شغلي ما بدهاش تفكير
\end{arabtext}} \\ 

\cellcolor{LimeGreen!30} \textbf{NP-1} & \cellcolor{LimeGreen!30} \RaggedRight{\begin{arabtext} \small  
أنا صار ليساعة بحكي وأنت بتقول لي بدي أفكر هاي شغلي ما بدهاش تفكير
\end{arabtext}} \\

\cellcolor{LimeGreen!30} \textbf{NP-2} & \cellcolor{LimeGreen!30} \RaggedRight{\begin{arabtext} \small 
أنا صار لي ساعة بحكي وأنت بتقول لي بدي أفكر هاي شغلي مابدهاش تفكير
\end{arabtext}} \\

\cellcolor{LimeGreen!30} \textbf{$\alpha$-Selected} & \cellcolor{LimeGreen!30} \RaggedRight{\begin{arabtext} \small 
أنا صار لي ساعة بحكي وأنت بتقول لي بدي أفكر هاي شغلة ما بدهاش تفكير
\end{arabtext}} \\ 

\cellcolor{LimeGreen!30} \textbf{$\alpha$-Rank} & \cellcolor{LimeGreen!30} \RaggedLeft{
\textbf{8}
} \\[0.2cm]

\\[0.1cm]
\toprule
\cellcolor{Green!30}
\textbf{\textit{Variety}}  & \cellcolor{Green!30} \RaggedLeft{\rule{0pt}{1.2em}\textbf{Palestine}} \\[0.2cm]  \toprule 
\cellcolor{LimeGreen!30} \textbf{Reference} & \cellcolor{LimeGreen!30} \RaggedRight{\begin{arabtext} \small  
خليكم، خليكم أصحاب أفعال مش أقوال
\end{arabtext}} \\  

\cellcolor{LimeGreen!30} \textbf{Rank-1} & \cellcolor{LimeGreen!30} \RaggedRight{\begin{arabtext} \small  
خليكم خليكم أصحاب أفعال ميش أقواي
\end{arabtext}} \\ 

\cellcolor{LimeGreen!30} \textbf{NP-1} & \cellcolor{LimeGreen!30} \RaggedRight{\begin{arabtext} \small  
خليكم خليكم أصحاب أفعال مش أقواي
\end{arabtext}} \\

\cellcolor{LimeGreen!30} \textbf{NP-2} & \cellcolor{LimeGreen!30} \RaggedRight{\begin{arabtext} \small 
خليكم خليكم أصحاب أفعال ميش أقوال
\end{arabtext}} \\

\cellcolor{LimeGreen!30} \textbf{$\alpha$-Selected} & \cellcolor{LimeGreen!30} \RaggedRight{\begin{arabtext} \small
خليكم خليكم أصحاب أفعال مش أقوال
\end{arabtext}} \\ 

\cellcolor{LimeGreen!30} \textbf{$\alpha$-Rank} & \cellcolor{LimeGreen!30} \RaggedLeft{
\textbf{10}
} \\[0.2cm]

\\[0.1cm]
\toprule
\cellcolor{Green!30}
\textbf{\textit{Variety}}  & \cellcolor{Green!30} \RaggedLeft{\rule{0pt}{1.2em}\textbf{UAE}} \\[0.2cm]  \toprule 
\cellcolor{LimeGreen!30} \textbf{Reference} & \cellcolor{LimeGreen!30} \RaggedRight{\begin{arabtext} \small  
ما صدقت يا بنتي انه في يوم من الأيام اشوف ولدي نايف معقوق فالسجن

\end{arabtext}} \\  

\cellcolor{LimeGreen!30} \textbf{Rank-1} & \cellcolor{LimeGreen!30} \RaggedRight{\begin{arabtext} \small  

ما صدقت يا بنتي إن في يوم من الأيام أشوف وليدي نايف معقوك في السجن

\end{arabtext}} \\ 

\cellcolor{LimeGreen!30} \textbf{NP-1} & \cellcolor{LimeGreen!30} \RaggedRight{\begin{arabtext} \small  

ما صدقت يا بنتي إن في يوم من الأيام أشوف وليدي نايف معقوق في السجن

\end{arabtext}} \\

\cellcolor{LimeGreen!30} \textbf{NP-2} & \cellcolor{LimeGreen!30} \RaggedRight{\begin{arabtext} \small 
ما صدقت يا بنتي إن في يوم من الأيام أشوف ولدي نايف معقوك في السجن
\end{arabtext}} \\

\cellcolor{LimeGreen!30} \textbf{$\alpha$-Selected} & \cellcolor{LimeGreen!30} \RaggedRight{\begin{arabtext} \small
ما صدقت يا بنتي إن في يوم من الأيام أشوف ولدي نايف معقوق في السجن
\end{arabtext}} \\ 

\cellcolor{LimeGreen!30} \textbf{$\alpha$-Rank} & \cellcolor{LimeGreen!30} \RaggedLeft{
\textbf{8}
} \\[0.2cm]

\\[0.1cm]
\toprule
\cellcolor{Green!30}
\textbf{\textit{Variety}}  & \cellcolor{Green!30} \RaggedLeft{\rule{0pt}{1.2em}\textbf{Algeria}} \\[0.2cm]  \toprule 
\cellcolor{LimeGreen!30} \textbf{Reference} & \cellcolor{LimeGreen!30} \RaggedRight{\begin{arabtext} \small  
ماتكسريش راسك مولاتي راني نستكلف بقول شي

\end{arabtext}} \\  

\cellcolor{LimeGreen!30} \textbf{Rank-1} & \cellcolor{LimeGreen!30} \RaggedRight{\begin{arabtext} \small  
ما تكصريش راصك مولتي راني نستكلف بكل شي

\end{arabtext}} \\ 

\cellcolor{LimeGreen!30} \textbf{NP-1} & \cellcolor{LimeGreen!30} \RaggedRight{\begin{arabtext} \small  

ما تكصريش راصك مولتي راني نستكلف بكل شي

\end{arabtext}} \\

\cellcolor{LimeGreen!30} \textbf{NP-2} & \cellcolor{LimeGreen!30} \RaggedRight{\begin{arabtext} \small 
ما تكصريش راسك مولتي راني نستكلف بكل شيء

\end{arabtext}} \\

\cellcolor{LimeGreen!30} \textbf{$\alpha$-Selected} & \cellcolor{LimeGreen!30} \RaggedRight{\begin{arabtext} \small
ما تكصريش راسك مولتي راني نستكلف بكل شي

\end{arabtext}} \\ 

\cellcolor{LimeGreen!30} \textbf{$\alpha$-Rank} & \cellcolor{LimeGreen!30} \RaggedLeft{
\textbf{3}
} \\[0.2cm]

\\[0.1cm]
\toprule
\cellcolor{Green!30}
\textbf{\textit{Variety}}  & \cellcolor{Green!30} \RaggedLeft{\rule{0pt}{1.2em}\textbf{Yemen}} \\[0.2cm]  \toprule 
\cellcolor{LimeGreen!30} \textbf{Reference} & \cellcolor{LimeGreen!30} \RaggedRight{\begin{arabtext} \small  

ليش بتفعل به هكذا؟

\end{arabtext}} \\  

\cellcolor{LimeGreen!30} \textbf{Rank-1} & \cellcolor{LimeGreen!30} \RaggedRight{\begin{arabtext} \small 

ليش متفعل فيه هكذا

\end{arabtext}} \\ 

\cellcolor{LimeGreen!30} \textbf{NP-1} & \cellcolor{LimeGreen!30} \RaggedRight{\begin{arabtext} \small

لش متفعل فيه هكذا

\end{arabtext}} \\

\cellcolor{LimeGreen!30} \textbf{NP-2} & \cellcolor{LimeGreen!30} \RaggedRight{\begin{arabtext} \small 

ليش متفعل فيه هكذا

\end{arabtext}} \\

\cellcolor{LimeGreen!30} \textbf{$\alpha$-Selected} & \cellcolor{LimeGreen!30} \RaggedRight{\begin{arabtext} \small
ليش بتفعل فيه هكذا
\end{arabtext}} \\ 

\cellcolor{LimeGreen!30} \textbf{$\alpha$-Rank} & \cellcolor{LimeGreen!30} \RaggedLeft{
\textbf{5}
} \\[0.2cm]

\\[0.1cm]
\toprule
\cellcolor{Green!30}
\textbf{\textit{Variety}}  & \cellcolor{Green!30} \RaggedLeft{\rule{0pt}{1.2em}\textbf{MSA - CV}} \\[0.2cm]  \toprule 
\cellcolor{LimeGreen!30} \textbf{Reference} & \cellcolor{LimeGreen!30} \RaggedRight{\begin{arabtext} \small  
تعبت من الرقود في السرير طوال اليوم

\end{arabtext}} \\  

\cellcolor{LimeGreen!30} \textbf{Rank-1} & \cellcolor{LimeGreen!30} \RaggedRight{\begin{arabtext} \small 

طعبت من الرقود في السرير طوال اليوم

\end{arabtext}} \\ 

\cellcolor{LimeGreen!30} \textbf{NP-1} & \cellcolor{LimeGreen!30} \RaggedRight{\begin{arabtext} \small

طعبت من الرقود في السرير طوال اليوم
\end{arabtext}} \\

\cellcolor{LimeGreen!30} \textbf{NP-2} & \cellcolor{LimeGreen!30} \RaggedRight{\begin{arabtext} \small 

تعبت من الركود في السرير طوال اليوم
\end{arabtext}} \\

\cellcolor{LimeGreen!30} \textbf{$\alpha$-Selected} & \cellcolor{LimeGreen!30} \RaggedRight{\begin{arabtext} \small

تعبت من الرقود في السرير طوال اليوم

\end{arabtext}} \\ 

\cellcolor{LimeGreen!30} \textbf{$\alpha$-Rank} & \cellcolor{LimeGreen!30} \RaggedLeft{
\textbf{3}
} \\[0.2cm]

\bottomrule
\end{tabular}
}
\caption{Qualitative examples across MSA and Arabic dialects where $\alpha$-interpolation recovers linguistically correct hypotheses missed by rank-1 and single-proxy selection by promoting lower-ranked beams. \textbf{Rank-1} denotes the top ASR hypothesis; \textbf{NP-1/NP-2}, nearest to Proxy~1/2; \textbf{$\alpha$-Selected}, interpolated choice; \textbf{$\alpha$-Rank}, beam rank; \textbf{CV}, Common Voice.
}

\label{tab:alpha-examples}
\end{table}


\end{document}